%% file: main.tex
\documentclass{article}

\usepackage{microtype}
\usepackage{graphicx}
\usepackage{booktabs} 

\usepackage{hyperref}

\usepackage[accepted]{mlsys2024}

\usepackage[ruled,algo2e]{algorithm2e}
\usepackage{textcomp}
\usepackage{xspace}
\usepackage{makecell}
\usepackage{float}

\usepackage{enumitem}

\usepackage{amsmath,amsthm}
\usepackage{amssymb}
\usepackage{soul}
\usepackage{subcaption}

\usepackage{pifont}
\newcommand{\cmark}{\ding{51}}%
\newcommand{\xmark}{\ding{55}}%

\usepackage{wrapfig}

\usepackage{ulem}

\usepackage[capitalize,noabbrev]{cleveref}
\usepackage{anyfontsize}
\newcommand{\tsref}[1]{\textsection\ref{#1}\xspace}
\usepackage{mdframed}

\theoremstyle{plain}
\newtheorem{theorem}{Theorem}[section]

\newtheorem{lemma}[theorem]{Lemma}

\theoremstyle{definition}
\newtheorem{definition}[theorem]{Definition}

\theoremstyle{remark}
\newtheorem{remark}[theorem]{Remark}

\newcommand\review[1]{
{\color{red}
{\textbf{Review:}
{\em#1}}}}
\newcommand\resolve[1]{
{\color{cyan}
{\textbf{Resolve:}
{\em#1}}}}

\newcommand{\switch}[1]{%
   \ifthenelse{\equal{#1}{0}}{\renewcommand{\review}[1]{}}{}
   \ifthenelse{\equal{#1}{0}}{\renewcommand{\resolve}[1]{}}{}}
\switch{1}
 \usepackage{comment}
\crefname{section}{\S}{\S\S}

\mlsystitlerunning{FedHE: An Efficient Homomorphic-Encryption-Based Privacy-Preserving Federated Learning System}

\begin{document}

\twocolumn[
\mlsystitle{FedML-HE: An Efficient Homomorphic-Encryption-Based Privacy-Preserving Federated Learning System}



\mlsyssetsymbol{equal}{*}

\begin{mlsysauthorlist}
\mlsysauthor{Weizhao Jin}{equal,usc}
\mlsysauthor{Yuhang Yao}{equal,cmu}
\mlsysauthor{Shanshan Han}{uci}
\mlsysauthor{Jiajun Gu}{cmu}
\mlsysauthor{Carlee Joe-Wong}{cmu}
\mlsysauthor{Srivatsan Ravi}{usc}
\mlsysauthor{Salman Avestimehr}{fedml}
\mlsysauthor{Chaoyang He}{fedml}
\end{mlsysauthorlist}

\mlsysaffiliation{usc}{University of Southern California}
\mlsysaffiliation{cmu}{Carnegie Mellon University}
\mlsysaffiliation{fedml}{FedML Inc.}
\mlsysaffiliation{uci}{University of California Irvine}

\mlsyscorrespondingauthor{Chaoyang He}{ch@fedml.ai}
\mlsyscorrespondingauthor{Weizhao Jin}{weizhaoj@usc.edu}
\mlsyscorrespondingauthor{Yuhang Yao}{yuhangya@andrew.cmu.edu}

\mlsyskeywords{Machine Learning, MLSys}

\vskip 0.3in

\input{abstract}
]

\printAffiliationsAndNotice{\mlsysEqualContribution} 


\input{intro}

\input{design}
\input{proofs}
\input{evaluation}

\input{background}
\input{conclusion}

\newpage
\bibliography{refs}
\bibliographystyle{mlsys2024}

\input{supp}

\end{document}

%% file: abstract.tex
\begin{abstract} 
Federated Learning trains machine learning models on distributed devices by aggregating local model updates instead of local data. However, privacy concerns arise as the aggregated local models on the server may reveal sensitive personal information by inversion attacks. Privacy-preserving methods, such as homomorphic encryption (HE), then become necessary for FL training. Despite HE's privacy advantages, its applications suffer from impractical overheads, especially for foundation models. In this paper, we present FedML-HE, \textit{the first practical federated learning system with efficient HE-based secure model aggregation}. FedML-HE proposes to selectively encrypt sensitive parameters, significantly reducing both computation and communication overheads during training while providing customizable privacy preservation. Our optimized system demonstrates considerable overhead reduction, particularly for large foundation models (e.g., $\sim$10x reduction for ResNet-50, and up to $\sim$40x reduction for BERT), demonstrating the potential for scalable HE-based FL deployment.

\end{abstract}

%% file: intro.tex
\section{Introduction}
\begin{figure}
    \centering
     \frame{\includegraphics[width=0.47\textwidth]{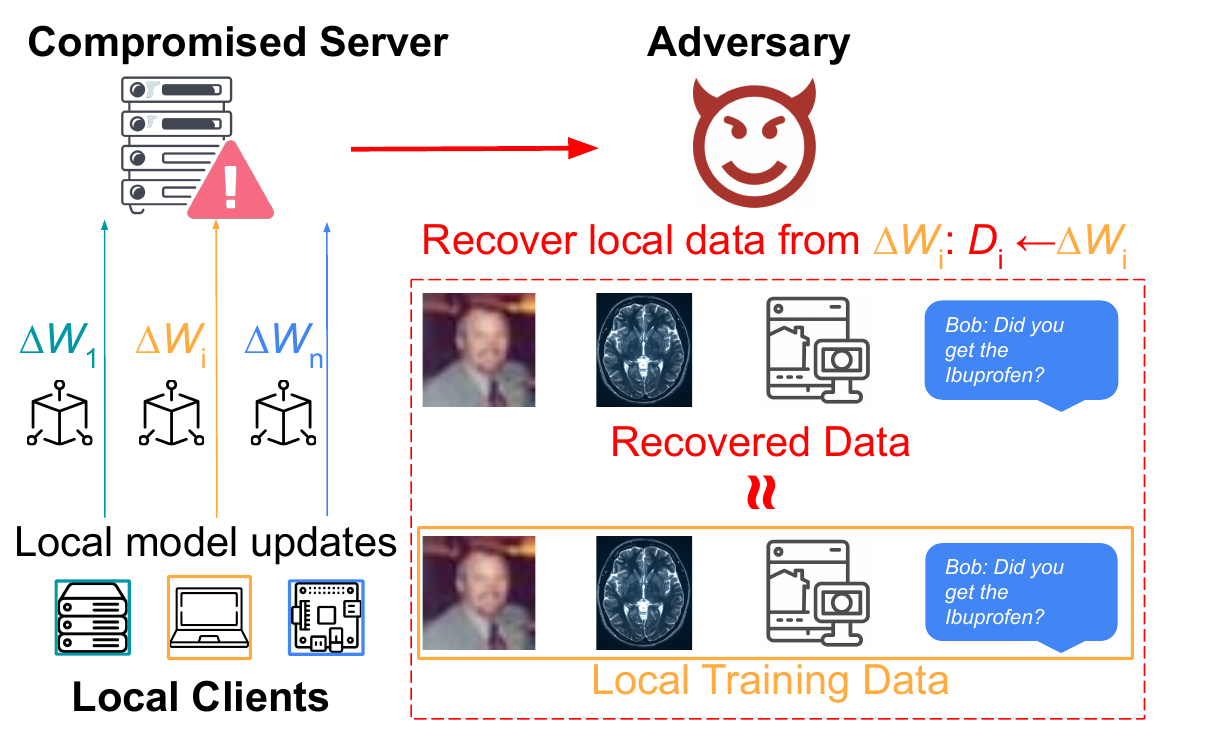}}
    \caption{Data Reconstruction Attacks: an adversarial server can recover local training data from local model updates.}
    \label{fig:attack_flow}
\end{figure}

Federated learning (FL) is increasingly popular in contemporary machine learning practices due to its ability to allow distributed clients to collectively train a global model without directly sharing data. Privacy preservation in standard federated learning systems depends on the distributed training process and the model aggregation function, such as  FedAvg~\cite{mcmahan2017communication}, FedSGD~\cite{fedsgd}, and 
FedGAN~\cite{rasouli2020fedgan}. 
In FL, 
instead of uploading raw data to a central server for training, clients train models locally and share their models with the server, where the local models are aggregated based on the aggregation functions. 
While FL ensures that local raw data do not leave their original locations, 
it remains vulnerable to eavesdroppers and malicious FL servers that might exploit plaintext local models (or model updates) 
to reconstruct sensitive training data, i.e., data reconstruction attacks or gradient inversion attacks in literature~\cite{zhu2019deep, criswell2014kcofi,bhowmick2018protection, hitaj2017deep,han2023fedmlsecurity,hatamizadeh2022gradvit,fowl2022decepticons}, as shown in Figure~\ref{fig:attack_flow}. 
This poses a privacy vulnerability especially when local models are trained on small local datasets, a common scenario in real-world applications such as smartphone text data for LLMs. Local models derived from these small datasets inherently contain fine-grained information, making it easier for adversaries to extract sensitive information from small model updates. 
\begin{table*}[t]
\centering
\begin{tabular}{|l|l|l|l|l|l|}
\hline
                        & \makecell{Model\\ Degradation} & Overheads & Client Dropout & \makecell{Interactive\\ Sync} & \makecell{Aggregated Model\\Visible To Server} \\ \hline
Differential Privacy    & With noise                   & \textbf{Light}        & \textbf{Robust}           & \textbf{No}     & Yes          \\ \hline
Secure Aggregation & \textbf{Exact}                       & Medium       & Susceptible            & Yes   & Yes           \\ \hline
Homomorphic Encryption  & \textbf{Exact}                & Large       & \textbf{Robust}          & \textbf{No}     &\textbf{No}         \\ \hline

\end{tabular}
\caption{Comparison of Differential Privacy, Secure Aggregation, and Homomorphic Encryption}
\label{tab:dp_mpc_he_compare}
\end{table*}

\begin{table*}[ht]
\centering
\begin{tabular}{|c|c|c|c|}
\hline
    Features&IBMFL& Nvidia FLARE & Ours \\ \hline
Homomorphic Encryption     &      \cmark &  \cmark  &   \cmark   \\ \hline
Threshold Key Management  &    \xmark &   \xmark &    \cmark  \\ \hline
Selective Parameter Encryption&      \xmark&  $\bigcirc$&   \cmark   \\ \hline 
Encrypted Foundation Model Training \ &     $\bigcirc$&  $\bigcirc$  &    \cmark  \\ \hline
\end{tabular}
\vspace{0.1cm}
\caption{Comparison with Existing HE-Based FL Systems. $\bigcirc$ implies limited support: for Selective Parameter Encryption, FLARE offers the (random) partial encryption option which does not have clear indications on privacy impacts; for Encrypted Foundation Model Training, the other two platforms require massive resources to train foundation models in encrypted federated learning.}
\label{tab:compare_intro}
\end{table*}

Existing defense methods that prevent privacy leakage from plaintext local models include differential privacy (DP)~\cite{truex2019hybrid,byrd2020differentially} and secure aggregation~\cite{bonawitz2017practical, so2022lightsecagg}. 
DP adds noise to original models but may result in model performance degradation due to the privacy noises introduced. On the other hand, secure aggregation employs zero-sum masks to shield local model updates, ensuring that the details of each update remain private. However, secure aggregation demands additional interactive synchronization steps and is sensitive to client dropout, making it less practical in real-world FL applications, where the unstable environments of clients face challenges such as unreliable internet connections, and software crashes.

As shown in Table~\ref{tab:dp_mpc_he_compare}, compared to the non-HE FL solutions above, homomorphic encryption (HE)~\cite{paillier1999public,gentry2009fully, cryptoeprint:2012/144, brakerski2014leveled, cheon2017homomorphic} offers a robust post-quantum secure solution that protects local models against attacks and \textit{provides stronger privacy guarantee while keeping the model aggregation with exact gradients}. 
HE-based federated learning (HE-FL) encrypts local models on clients and performs model aggregation over ciphertexts on the server. This approach enables secure federated learning deployments with exactly the same model performance as vanilla FL 
and has been adopted by several FL systems~\cite{roth2022nvidia,ibmfl,zhang2020batchcrypt, du2023efficient} and a few domain-specific applications~\cite{stripelis2021secure, yao2022fedgcn}. 

\begin{figure}[ht]
\includegraphics[width=0.49\textwidth]{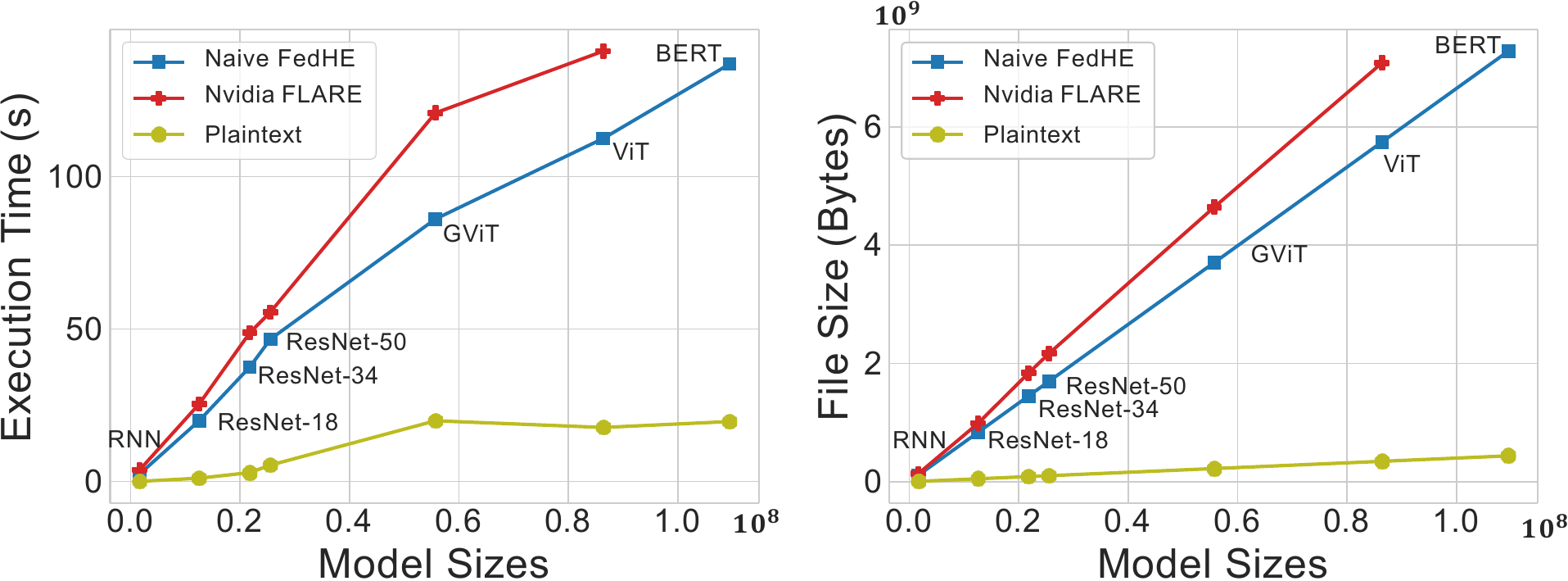}
\caption{Computational (left) and Computation (right) Overhead Comparison for Models of Different Sizes: Naive FedML-HE vs. Nvidia FLARE  vs. Plaintext Aggregation. Due to TenSeal's larger file sizes, FLARE did not finish the run on BERT on our 32GB memory machine.} 
\label{fig:comp}
\end{figure}

Despite the advantages of homomorphic encryption, HE remains a powerful but complex cryptographic foundation with impractical overheads (as shown in Figure~\ref{fig:comp}) for most real-world applications. Prior FL-HE solutions mainly employ existing generic HE methods without sufficient optimization for large-scale FL deployment~\cite{roth2022nvidia,ibmfl,zhang2020batchcrypt, du2023efficient}. The scalability of encrypted computation and communication during federated training then becomes a bottleneck, restricting its feasibility for real-world scenarios. This HE overhead limitation is particularly noticeable (\textit{commonly $\sim$15x increase in both computation and communication}~\cite{gouert2022new}) when training large foundation models across resource-constrained devices, where encrypted computing and communication of large models might take considerably longer than the actual model training.
\label{sec:microbenchmark}
\noindent It is widely known that HE inevitably introduces large overheads regarding both computation and communication~\cite{gouert2022new}. To verify this, we evaluate the vanilla HE implementation to pinpoint the overhead bottlenecks.

\textit{Observation}: As shown by the evaluation results in Figure~\ref{fig:comp}, the computational and communication (package size) overheads introduced by HE is $O(n)$, both growing linearly with the input size $n$, which in our case the sizes of the models for aggregation. \emph{Although the unoptimized system is faster than Nvidia FLARE, the execution time and file size are still impractical, especially for large models.}


\begin{figure*}[ht]
\centering
\includegraphics[width=1\textwidth]{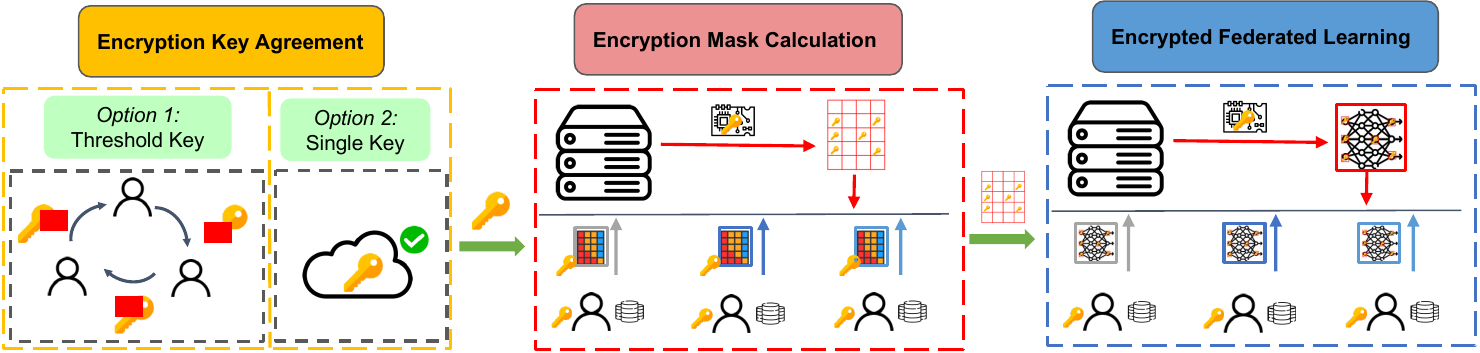}
\caption{FedML-HE System Pipeline: in the \textbf{Encryption Key Agreement} stage, clients can either use distributed threshold key agreement protocol or outsource a trusted key authority. We simplify the illustration here by abstracting the key pair of the public key and secret key (partial secret keys if using threshold protocol) as one key; in the \textbf{Encryption Mask Calculation} stage, clients use local datasets to calculate local model sensitivity maps which are homomorphically aggregated at the server to generate an encryption mask; in the \textbf{Encrypted Federated Learning} stage, clients use homomorphic encryption with encryption mask to protect local model updates where the server aggregates them but does not have access to sensitive local models.}
\label{fig:mlops-flow}
\end{figure*}

To address these challenges, we propose FedML-HE, an efficient Homomorphic Encryption-based privacy-preserving FL system with \textit{Selective Parameter Encryption}, designed for practical deployment across distributed edge devices. Our system significantly reduces communication and computation overheads, enabling HE-based federated learning to be more accessible and efficient in real-world scenarios (comparison with other popular HE-based FL work can be found in Table~\ref{tab:compare_intro}).


\noindent\textbf{Key contributions}: 
\begin{itemize}[noitemsep,topsep=0pt]

    \item We propose FedML-HE, the first practical Homomorphic Encryption-based privacy-preserving FL system that supports encryption key management, encrypted FL platform deployment, encryption optimizations to reduce overhead, and is designed to support efficient foundation model federated training.
    \item We propose \textbf{Selective Parameter Encryption} that selectively encrypts the most privacy-sensitive parameters to minimize the size of encrypted model updates while providing customizable privacy 
    preservation. 
    \item Theoretical privacy analysis shows the HE system can ensure privacy under single-key and threshold adversaries and encrypting most sensitivity parameters provides orders-of-magnitude better privacy guarantees. 
    
    \item  Extensive experiments show that the optimized system achieves significant overhead reduction while preserving privacy against state-of-the-art ML privacy attacks, particularly for large models (e.g., $\sim$10x reduction for HE-federated training ResNet-50 and up to $\sim$40x reduction for BERT), demonstrating the potential for real-world HE-based FL deployments.
\end{itemize}

%% file: design.tex
\section{FedML-HE System Design}
\label{sec:system}


In this section, we first provide the overview of FedML-HE system in \tsref{sec:system_overview}, define the threat model in \tsref{sec:threat_model}, describe the algorithmic design of FedML-HE in \tsref{sec:algo}, propose our efficient optimization method \textbf{Selective Parameter Encryption} after pinpointing the overhead bottleneck in \tsref{sec:opt}, and explain how we integrate homomorphic encryption in federated learning from a software framework perspective in \tsref{sec:he_fedml}. 

\subsection{System Overview}
\label{sec:system_overview}

As shown in Figure~\ref{fig:mlops-flow}, our efficient HE-based federated training process at a high level goes through three major stages: \textit{(1)} Encryption key agreement: the clients either use threshold HE key agreement protocol or trusted key authority to generate HE keys; \textit{(2)} Encryption mask calculation: the clients and the server apply \textbf{Selective Parameter Encryption} method using homomorphic encryption to agree on a selective encryption mask; \textit{(3)} Encrypted federated learning: the clients selectively encrypt local model updates using the homomorphic encryption key and the encryption mask for efficient privacy-preserving training. 

\subsection{Threat Model}
\label{sec:threat_model}

We define a semi-honest adversary $\mathcal{A}$ that can corrupt the aggregation server or any subset of local clients. $\mathcal{A}$ follows the protocol but tries to learn as much information as possible. Loosely speaking, under such an adversary, the security definition requires that only the private information in local models from the corrupted clients will be learned when $\mathcal{A}$ corrupts a subset of clients; no private information from local models nor global models will be learned by $\mathcal{A}$ when $\mathcal{A}$ corrupts the aggregation server.

When $\mathcal{A}$ corrupts both the aggregation server and a number of clients, the default setup where the private key is shared with all clients (also with corrupted clients) will allow $\mathcal{A}$ to decrypt local models from benign clients (by combining encrypted local models received by the corrupted server and the private key received by any corrupted client). This issue can be mitigated by adopting the threshold or multi-key variant of HE where decryption must be collaboratively performed by a certain number of clients~\cite{aloufi2021computing, ma2022privacy, du2023efficient}. Since the multi-party homomorphic encryption issue is not the focus of this work, in the rest of the paper we default to a single-key homomorphic encryption setup, but details on threshold homomorphic encryption federated learning setup and microbenchmarks are provided in the appendix.


\subsection{Algorithm for HE-Based Federated Aggregation}
\label{sec:algo}


\noindent Privacy-preserving federated learning systems utilize homomorphic encryption to enable the aggregation server to combine local model parameters without viewing them in their unencrypted form by designing homomorphic encrypted aggregation functions. We primarily focus on FedAvg~\cite{mcmahan2017communication}, which has been proved as still one of the most robust federated aggregation strategies while maintaining computational simplicity~\cite{wang2022unreasonable}.


Our HE-based secure aggregation algorithm, as illustrated in Algorithm~\ref{alg:fedml-fhe}, can be summarized as: given an aggregation server and $N$ clients, each client $i\in [N]$ owns a local dataset $\mathcal{D}_i$ and initializes a local model $\mathbf{W}_i$ with the aggregation weighing factor $\alpha_i$; the key authority or the distributed threshold key agreement protocol generates a key pair $(pk, sk)$ and the crypto context, then distributes it to clients and server (except the server only gets the crypto context which is public configuration). The clients and the server then collectively calculate the encryption mask $\mathbf{M}$ for \textbf{ Selective Parameter Encryption} also using homomorphic encryption.
At every communication round $t \in [T]$, the server performs the aggregation 
$$[\mathbf{W}_{\text {glob}}] =\sum_{i=1}^N \alpha_i [\![\mathbf{M} \odot \mathbf{W}_i]\!] + \sum_{i=1}^N \alpha_i ((\mathbf{1}-\mathbf{M})\odot \mathbf{W}_i),$$ 
where $[\mathbf{W}_{\text {glob}}]$ is the partially-encrypted global model, $\mathbf{W}_i$ is the $i$-th plaintext local model where $[\![]\!]$ indicates the portion of the model that is fully encrypted, $\alpha_i$ is the aggregation weight for client $i$, and $\mathbf{M}$ is the model encryption mask.

Note that the aggregation weights can be either encrypted or in plaintext depending on whether the aggregation server is trustworthy enough to obtain that information. In our system, we set the aggregation weights to be plaintext by default. We only need one multiplicative depth 
of HE multiplication in our algorithm for weighting, which is preferred to reduce HE multiplication operations. Our system can also be easily extended to support more FL aggregation functions with HE by encrypting and computing the new parameters in these algorithms (e.g. FedProx~\cite{li2020federated}). Additionally, in Algorithm~\ref{alg:fedml-fhe}, optional local differential privacy noise can be easily added after local models are trained if there is an extra desire for differential privacy. 

\begin{algorithm}[ht!]
\SetKwFor{ForPar}{for}{do in parallel}{end forpar}
    \caption{HE-Based Federated Aggregation}
    \label{alg:fedml-fhe}
    \begin{itemize}[itemsep=0em]
        \item $[\![\mathbf{W}]\!]$: the fully encrypted model $|$ $[\mathbf{W}]$: the partially encrypted model;
        \item $p$: the ratio of parameters for selective encryption;
        \item $b$: (optional) differential privacy parameter.
    \end{itemize}
\tcp{Key Authority Generate Key}
$(pk, sk) \gets HE.KeyGen(\lambda)$;

\tcp{Local Sensitivity Map Calculation}
\ForPar{each client $i \in [N]$}{
    $\mathbf{W}_i \gets Init(\mathbf{W})$;

    $\mathbf{S}_i \gets Sensitivity(\mathbf{W}, \mathcal{D}_i)$;
    
    $[\![\mathbf{S}_i]\!] \gets Enc(pk, \mathbf{S}_i)$;
    
    Send $[\![\mathbf{S}_i]\!]$ to server;
}
\tcp{Server Encryption Mask Aggregation}
$[\![\mathbf{M}]\!] \gets Select(\sum_{i=1}^N \alpha_i [\![\mathbf{S}_i]\!], p$);

\tcp{Training}
\raggedright 
\For{$t = 1, 2, \dots, T$}{
    \ForPar{each client $i \in [N]$}{
        \If{$t = 1$}{
             Receive $[\![\mathbf{M}]\!]$ from server;\\
            $\mathbf{M} \gets HE.Dec(sk,  [\![\mathbf{M}]\!])$;\\
        }
        \If{$t > 1$}{
            Receive $[\mathbf{W}_\text{glob}]$ from server;\\
            $\mathbf{W}_i \gets HE.Dec(sk, \mathbf{M} \odot [\mathbf{W}_\text{glob}]) + (\mathbf{1}-\mathbf{M})\odot [\mathbf{W}_\text{glob}]$;\\
        }
        $\mathbf{W}_i \gets Train(\mathbf{W}_i, \mathcal{D}_i)$;\\
        \tcp{Additional Differential Privacy}
        \If{Add DP}{
        $\mathbf{W}_i \gets \mathbf{W}_i + Noise(b)$;
        }
        $[\mathbf{W}_i] \gets HE.Enc(pk, \mathbf{M} \odot \mathbf{W}_i) + (\mathbf{1}-\mathbf{M})\odot \mathbf{W}_i$;\\
        Send $[\mathbf{W}_i]$ to server $\mathcal{S}$;\\
    }
    \tcp{Server Model Aggregation}
    $[\mathbf{W}_{\text {glob}}] \gets \sum_{i=1}^N \alpha_i [\![\mathbf{M} \odot \mathbf{W}_i]\!] + \sum_{i=1}^N \alpha_i ((\mathbf{1}-\mathbf{M})\odot \mathbf{W}_i)$;\\
}
\end{algorithm}

We will explain in detail how the encryption mask $\mathbf{M}$ is formalized in \tsref{sec:opt}.

\input{opt}

\subsection{Software Framework: Homomorphic Encryption In Federated Learning}
\label{sec:he_fedml}
\noindent In this part, we will illustrate how we design our HE-based aggregation from a software framework perspective. 
\begin{figure}[ht!]
  \centering
 \includegraphics[width=0.48\textwidth]{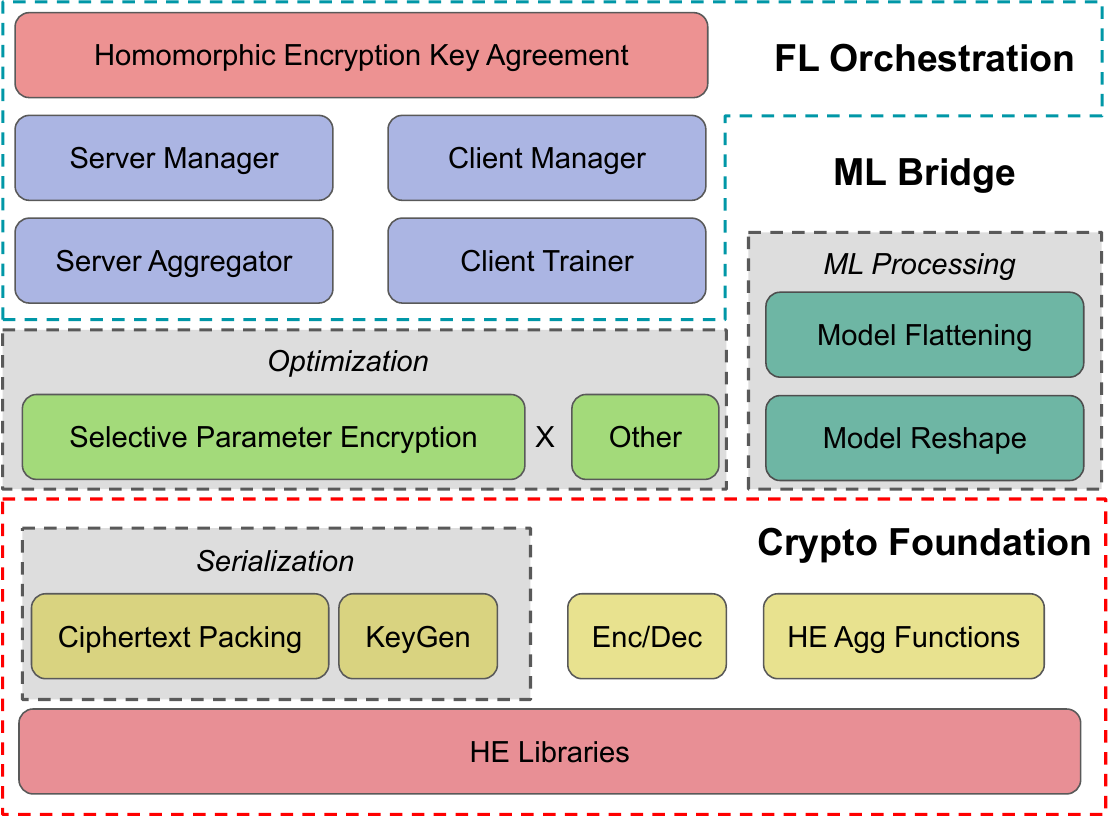}
  \caption{Framework Structure: our framework consists of a three-layer structure including Crypto Foundation to support basic HE building blocks, ML Bridge to connect crypto tools with ML functions, and FL Orchestration to coordinate different parties during a task. }
    \label{fig:design}
\end{figure}

Figure~\ref{fig:design} provides a high-level design of our framework, which consists of three major layers: 
\begin{itemize}[noitemsep,topsep=0pt]
    \item \textbf{Crypto Foundation.} The foundation layer is where Python wrappers are built to realize HE functions including key generation, encryption/decryption, secure aggregation, and ciphertext serialization using open-sourced HE libraries;
    \item \textbf{ML Bridge.} The bridging layer connects the FL system orchestration and cryptographic functions. Specifically, we have ML processing APIs to process inputs to HE functions from local training processes and outputs vice versa. Additionally, we realize the optimization module here to mitigate the HE overheads;
    \item \textbf{FL Orchestration.} The FL system layer is where the key authority server manages the key distribution and the (server/client) managers and task executors orchestrate participants.
\end{itemize}
Our layered design makes the HE crypto foundation and the optimization module \textit{semi-independent}, allowing different HE libraries to be easily switched into FedML-HE and further FL optimization techniques to be easily added to the system.

%% file: opt.tex
\subsection{Efficient Optimization by Selective Parameter Encryption}
\label{sec:opt}
\begin{figure*}[ht]
\includegraphics[width=1\textwidth]{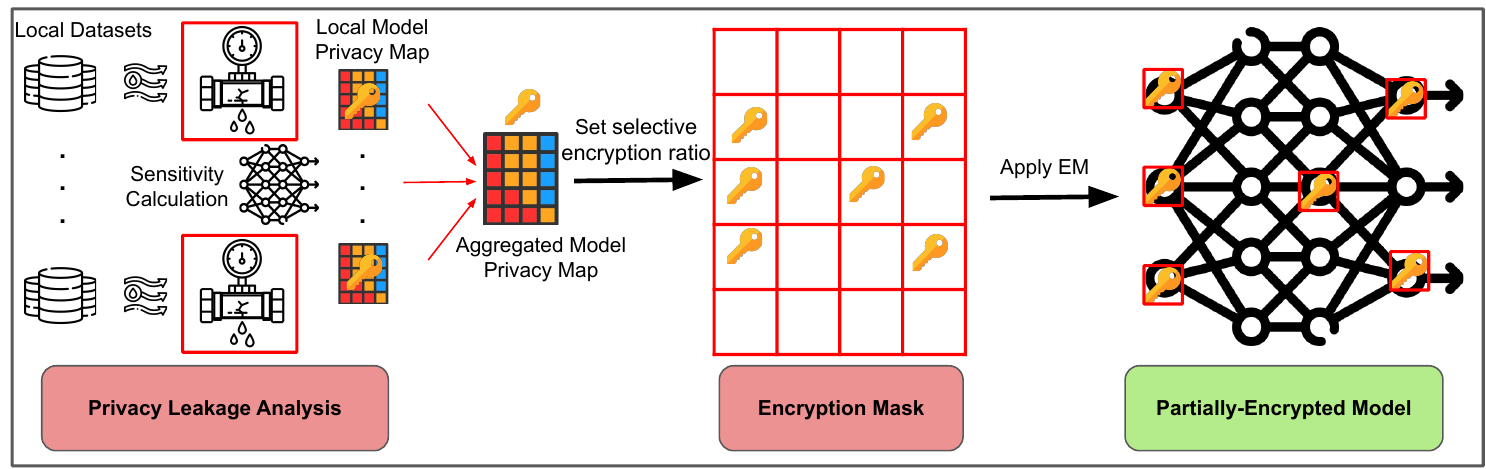}
\caption{\textbf{Selective Parameter Encryption}: in the initialization stage, clients first calculate privacy sensitivities on the model using its own dataset and local sensitivities will be securely aggregated to a global model privacy map. The encryption mask will be then determined by the privacy map and a set selection value $p$ per overhead requirements and privacy guarantee. Only the masked parameters will be aggregated in the encrypted form.}
\label{fig:mask}
\end{figure*}

\begin{figure*}[ht]
\includegraphics[width=1.0\textwidth]{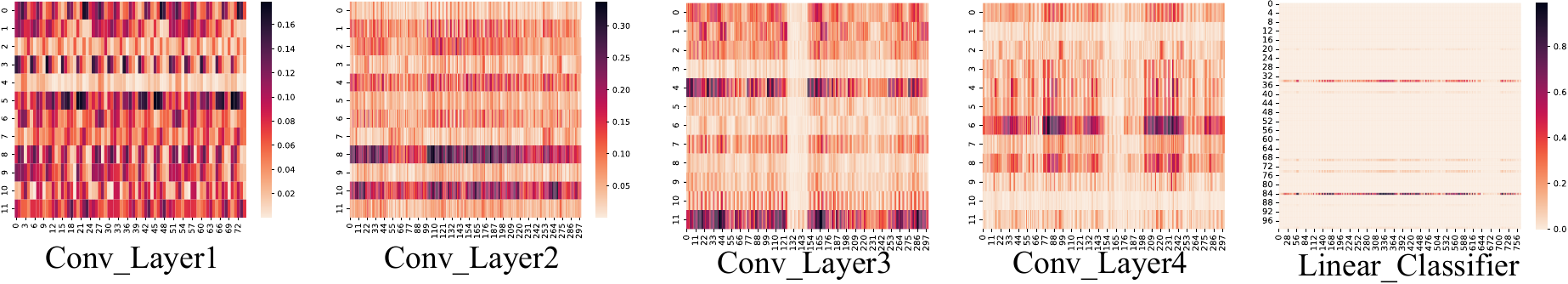}
\caption{Model Privacy Map Calculated by Sensitivity on LeNet: darker color indicates higher sensitivity. Each subfigure shows the sensitivity of parameters of the current layer. The sensitivity of parameters is imbalanced and many parameters have very little sensitivity (its gradient is hard to be affected by tuning the data input for attack).}
\label{fig:lenet_map}
\end{figure*}


\noindent Fully encrypted models can guarantee no access to plaintext local models from the adversary with high overheads. However, previous work on privacy leakage analysis shows that ``partial transparency'', e.g. hiding parts of the models~\citep{hatamizadeh2022gradvit, mo2020layer}, can limit an adversary's ability to successfully perform attacks like gradient inversion attacks~\citep{lu2022april}. We therefore propose \textbf{Selective Parameter Encryption} to \textit{selectively encrypt the most privacy-sensitive parameters} in order to reduce impractical overhead while providing customizable privacy preservation; see Figure~\ref{fig:mask}.  



\textbf{Step 1: Privacy Leakage Analysis on Clients.} Directly performing a gradient inversion attack~\cite{wei2020framework} and evaluating the success rate of the attack can take much more time than the model training. 
We then adopt sensitivity~\cite{novaksensitivity, sokolic2017robust, mo2020layer} for measuring the general privacy risk on gradients w.r.t. input. 
Given model $\mathbf{W}$ and $K$ data samples with input matrix $\mathbf{X}$ and ground truth label vector $\mathbf{y}$, we compute the sensitivity for each parameter $w_m$ by 
$\frac{1}{K} \sum_{k=1}^K \left\|J_m\left(y_k\right) \right\|,$ 
where $J_m\left(y_k\right) =\frac{\partial}{\partial y_k}\left(\frac{\partial \ell\left(\mathbf{X}, \mathbf{y}, \mathbf{W}\right)}{\partial w_m}\right) \in R$, $\ell(\cdot)$ is the loss function given $\mathbf{X}$, $\mathbf{y}$ and $\mathbf{W}$, and $\left\|\cdot \right\|$ calculates the absolute value. The intuition is to calculate how large the gradient of the parameter will change with the true output $y_k$ for each data point $k$. Each client $i$ then sends the encrypted parameters sensitivity matrix $[\![\mathbf{S}_i]\!]$ to the server.




As shown in Figure~\ref{fig:lenet_map}, different parts of a model contribute to attacks by revealing uneven amounts of information. Using this insight, we propose to only select and encrypt parts of the model that are more important and susceptible to attacks to reduce HE overheads while preserving adequate privacy. 

\textbf{Step 2: Encryption Mask Agreement across Clients.} The sensitivity map is dependent on the data it is processed on. With potentially heterogeneous data distributions, the server aggregates local sensitivity maps to a global privacy map $\sum_{i=1}^N \alpha_i [\![\mathbf{S}_i]\!]$. The global encryption mask $\mathbf{M}$ is then configured using a privacy-overhead ratio $p\in[0,1]$ which is the ratio of selecting the most sensitive parameters for encryption. The global encryption mask is then shared among clients as part of the federated learning configuration.

%% file: proofs.tex
\section{Privacy By Selective Parameter Encryption}
In this section, we first provide proof to analyze the privacy of fully encrypted federated learning and then analyze the privacy guarantee of Selective Parameter Encryption.
\subsection{Proof of Base Protocol}
In this subsection, we prove the privacy of base protocol where homomorphic-encryption-based federated learning utilizes the full model parameter encryption (i.e., the selective parameter encryption rate is set to be \textit{1}). We define the adversary in Definition~\ref{def:adv} and privacy in Definition~\ref{def:privacy}. 

\begin{definition}[Single-Key Adversary]
\label{def:adv}
\textit{A semi-honest adversary $\mathcal{A}$ can corrupt (at the same time) any subset of $n$ learners and the aggregation server, but not at the same time.}
\end{definition}

Note that the ref of the proof assumes the single-key setup and the privacy of the threshold variant of HE-FL (as shown in Definition~\ref{def:adv-threshold}) can be easily proved by extending the proofs of threshold homomorphic encryption~\cite{boneh2006chosen, laud2008threshold, asharov2012multiparty}.

\begin{definition}[Threshold Adversary]
\label{def:adv-threshold}
\textit{ A semi-honest adversary $\mathcal{A_Th}$ can corrupt (at the same time) any subset of $n-k$ learners and the aggregation server.}
\end{definition}

\begin{definition}[Privacy]
    \label{def:privacy}
	\textit{A homomorphic-encryption federated learning protocol $\pi$ is simulation secure in the presence of a semi-honest adversary $\mathcal{A}$, there exists a simulator $\mathcal{S}$  in the ideal world that also corrupts the same set of parties and produces an output identically distributed to $\mathcal{A}$'s output in the real world.}
\end{definition}

\noindent\textbf{Ideal World.} Our ideal world functionality $\mathcal{F}$  interacts with
learners and the aggregation server as follows:

\begin{itemize}[leftmargin=*,itemsep=0pt,topsep=0pt]
    \item Each learner sends a registration message to $\mathcal{F}$ for a federated training model task $\mathbf{W}_\text{glob}$. $\mathcal{F}$  determines a subset $N' \subset N$ of learners whose data can be used to compute the global model $\mathbf{W}_\text{glob}$. 
    
    \item Both honest and corrupted learners upload their local models to $\mathcal{F}$.
    

    \item If local models $\vec{\mathbf{W}}$ of learners in $N'$ are enough to compute
        $\mathbf{W}_\text{glob}$, $\mathcal{F}$  sends $\mathbf{W}_\text{glob} \gets \sum_{i=1}^{N'} \alpha_i \mathbf{W}_i$ to all learners in $N'$, otherwise $\mathcal{F}$  sends empty message $\bot$.
\end{itemize}

\noindent\textbf{Real World.} In real world, $\mathcal{F}$  is replaced by our
protocol described in Algorithm~\ref{alg:fedml-fhe} with full model parameter encryption.

We describe a simulator $\mathcal{S}$  that simulates the view of the $\mathcal{A}$  in the real-world execution of our protocol. Our privacy definition~\ref{def:privacy} and the simulator $\mathcal{S}$ prove both confidentiality and correctness. We omit the simulation of the view of $\mathcal{A}$ that corrupts the aggregation server here since the learners will not receive the ciphertexts of other learners' local models in the execution of $\pi$ thus such a simulation is immediate and trivial.

\input{simulator}

\subsection{Proof of Encrypted Learning by DP Theory}

\begin{definition}[Adjacent Datasets]
\label{def:adj_datasets}
Two datasets $D_1$ and $D_2$ are said to be adjacent if they differ in the data of exactly one individual. Formally, they are adjacent if:
$$
\left|D_1 \Delta D_2\right|=1
$$
\end{definition} 

\begin{definition}[$\epsilon$-Differential Privacy]
\label{def:dp}
A randomized algorithm $\mathcal{M}$ satisfies $\epsilon$-differential privacy if for any two adjacent datasets $D_1$ and $D_2$, and for any possible output $O \subseteq \operatorname{Range}(\mathcal{F})$, the following inequality holds:
$$
\frac{\operatorname{Pr}\left[\mathcal{M}\left(D_1\right) \in O\right]}{\operatorname{Pr}\left[\mathcal{M}\left(D_2\right) \in O\right]} \leq e^\epsilon
$$
Smaller values of the privacy parameter $\epsilon$ imply stronger privacy guarantees.
\end{definition} 

\begin{definition}[Laplace mechanism]
\label{def:laplace}
Given a function $f: \mathcal{D} \rightarrow \mathbb{R}$, 

where $\mathcal{D}$ is the domain of the dataset and $d$ is the dimension of the output, the Laplace mechanism adds Laplace noise to the output of $f$.

Let $b$ be the scale parameter of the Laplace distribution, which is given by:
$$
\operatorname{Lap}(x \mid b)=\frac{1}{2 b} e^{-\frac{|x|}{b}}
$$

Given a dataset $D$, the Laplace mechanism $\mathcal{F}$ is defined as:
$$
\mathcal{M}(D)=f(D)+\operatorname{Lap}(0 \mid b)^d
$$
\end{definition} 

\begin{definition}[Sensitivity]
\label{def:sensitivity}
To ensure $\epsilon$-differential privacy, we need to determine the appropriate scale parameter $b$. This is where the sensitivity of the function $f$ comes into play. The sensitivity $\Delta f$ of a function $f$ is the maximum difference in the output of $f$ when applied to any two adjacent datasets:
$$
\Delta f=\max _{D_1, D_2:\left|D_1 \Delta D_2\right|=1}\left\|f\left(D_1\right)-f\left(D_2\right)\right\|_1
$$
\end{definition} 
Based on Definition~\ref{def:adj_datasets}, \ref{def:dp}, \ref{def:laplace} and \ref{def:sensitivity} we have

\begin{lemma}[Achieving $\epsilon$-Differential Privacy by Laplace Mechanism~\cite{dwork2008differential, abadi2016deep}]
\label{lemma:dp_laplace}
To achieve $\epsilon$-differential privacy, we choose the scale parameter $b$ as:
$$
b=\frac{\Delta f}{\epsilon}
$$

With this choice of $b$, the Laplace mechanism $\mathcal{F}$ satisfies $\epsilon$-differential privacy.

\end{lemma}

By adding noise $\operatorname{Lap}(0 \mid b)^d$ on one parameter in the model gradient where $b=\frac{\Delta f}{\epsilon}$, we can achieve $\epsilon$-differential privacy. We then show homomorphic encryption provides a much stronger differential privacy guarantee.

\begin{theorem}[Achieving $0$-Differential Privacy by Homomorphic Encryption]
\label{theorem:he_dp}
For any two adjacent datasets $D_1$ and $D_2$, since $\mathcal{M}(D)$ is computationally indistinguishable, we have
$$
\frac{\operatorname{Pr}\left[\mathcal{M}\left(D_1\right) \in O\right]}{\operatorname{Pr}\left[\mathcal{M}\left(D_2\right) \in O\right]} \leq e^{\epsilon}.
$$
We then have $\epsilon = 0$ if $O$ is encrypted. 
\end{theorem}

In other words, $\mathcal{A}$ cannot retrieve sensitive information from encrypted parameters.

\subsection{Proof of Selective Parameter Selection}

\begin{lemma}[Sequential Composition~\cite{dwork2008differential},]
\label{lemma:sequential}
If $\mathcal{M}_1(x)$ satisfies $\epsilon_1$-differential privacy and $\mathcal{M}_2(x)$ satisfies $\epsilon_2$-differential privacy, then the mechanism $\mathcal{G}(x)=\left(\mathcal{M}_1(x), \mathcal{M}_2(x)\right)$ which releases both results satisfies $(\epsilon_1+\epsilon_{2})$-differential privacy
\end{lemma}


Based on Lemma~\ref{lemma:dp_laplace},~\ref{lemma:sequential} and Thoerem~\ref{theorem:he_dp}, we can now analyze the privacy of Selective Parameter Encryption

\begin{theorem}[Achieving $\sum_{i\in {[N] / \mathcal{S}}} \frac{\Delta f_i}{b}$-Differential Privacy by Partial Encryption]
If we apply Homomorphic Encryption on partial model parameters $\mathcal{S}$ and Laplace Mechanism on remaining model parameters $[N] / \mathcal{S}$ with fixed noise scale $b$. For each parameter $i\in [N] / \mathcal{S}$, we have $\epsilon_i = \frac{\Delta f_i}{b}$. Such partial encryption satisfies $\sum_{i\in {[N] / \mathcal{S}}} \frac{\Delta f_i}{b}$-differential privacy.

\end{theorem}
Let $J = \sum_{i=1}^N \frac{\Delta f_i}{b}$ and assume $\Delta f \sim \mathcal{U}(0,1)$ where $\mathcal{U}$ represents the uniform distribution, we can then show the privacy cost of adding Laplace noise on all parameters, random parameter encryption, and selective parameter encryption.
\begin{remark}[Achieving $J$-Differential Privacy by Laplace Mechanism on All Model Parameters]
If we add Laplace noise on all parameters with fixed noise scale $b$, it satisfies $J$-differential privacy.
\end{remark}

\begin{remark}[Achieving $(1-p) J$-Differential Privacy by Random Selection]
If we randomly select model parameters with probability $p$ and homomorphically encrypt the remaining parameters, it satisfies $(1-p) J$-differential privacy.

\end{remark}

\begin{remark}[Achieving $(1-p)^2 J$-Differential Privacy by Sensitive Parameter Selection]
If we select the most sensitive parameters with ratio $p$ and homomorphically encrypt the remaining parameters, it satisfies $(1-p)^2 J$-differential privacy.
\end{remark}

\textbf{Key Observation}: Selective Parameter Encryption requires \textit{$(1-p)^2$ times less privacy budget} than random selection and complete differential privacy with the same privacy preservation.

%% file: simulator.tex
\noindent\textbf{Simulator.} In the ideal world, $\mathcal{S}$ receives $\lambda$ and $1^n$ from $\mathcal{F}$ and executes the following steps:

\begin{enumerate}[itemsep=0mm]
    \item $\mathcal{S}$ chooses a uniformly distributed random tape $r$.
    \item $\mathcal{S}$ runs the key generation function to sample $pk$: $(pk, sk) \leftarrow \mathit{HE}.KeyGen(\lambda)$.
    \item For a chosen $i$th learner, $\mathcal{S}$ runs the encryption function to sample: $(c_i) \leftarrow \mathit{HE}.Enc(pk, r^{\left|\mathbf{W}_i\right|})$.
    \item $\mathcal{S}$ repeats Step $3$ for all other learners to obtain $\vec{c}$, and runs the federated aggregation function $f$ to sample: $(c_\text{glob}) \leftarrow \mathit{HE}.Eval(\vec{\mathbf{c}}, f)$.
\end{enumerate}

The execution of $\mathcal{S}$ implies that:

$\left\{\left(c_i, c_\text{glob}\right)\right\} \stackrel{\mathrm{s}}{\equiv}\left\{\left(\mathit{HE}.Enc(pk, \mathbf{W}_i), \mathit{HE}.Eval(\vec{\mathbf{W}}, f)\right)\right\}$

Thus, we conclude that $\mathcal{S}$'s output in the ideal world is computationally indistinguishable from the view of $\mathcal{A}$ in a real world execution:

$\left\{\mathcal{S}\left(1^n,\left(\lambda\right)\right)\right\} \stackrel{\mathrm{s}}{\equiv}\left\{\operatorname{view}^\pi\left(\lambda\right)\right\}$,

where $\operatorname{view}$ is the view of $\mathcal{A}$ in the real execution of $\pi$.

%% file: evaluation.tex
\section{Evaluation}

\noindent In this section, we focus on the evaluation results to show how our proposed universal optimization scheme largely mitigates these overheads for real-world deployment but still guarantees adequate defense against privacy attacks. Note that additional experimental results regarding other FL system aspects are included in in the appendix.

\subsection{Experiment Setup}
\noindent\textbf{Models.} We test our framework on models in different ML domains with different sizes including Llama-2 (7 billion) (more details in in the appendix).


\noindent\textbf{HE Libraries.} We implement our HE core using both PALISADE and TenSEAL. Unless otherwise specified, our results show the evaluation of the PALISADE version. 

\noindent\textbf{Default Crypto Parameters.} Unless otherwise specified, we choose the multiplicative depth of 1, the scaling factor bit digit of 52, an HE packing batch size of 4096, and a security level of 128 as our default HE cryptographic parameters during the evaluation.

\noindent\textbf{Microbenchmark.} For microbenchmarking HE overheads, we use an Intel 8-core 3.60GHz i7-7700 CPU with 32 GB memory and an NVIDIA Tesla T4 GPU on Ubuntu 18.04.6. 




\subsection{Optimizations}
\label{sec:opt_results}

To mitigate the HE overhead surge, our optimization scheme \textbf{Selective Parameter Encryption} works by selecting sensitive portions of parameters for encrypted computation while leaving the rest in plaintext per desired overhead expectations and privacy promise. In this section, we first evaluate the overhead optimization from \textbf{Selective Parameter Encryption} and then use the state-of-the-art privacy attacks to evaluate the effectiveness of our selection defense during FL training.

Note that other parameter efficiency techniques~\cite{tang2019doublesqueeze, hu2021lora} for both training-from-scratch and fine-tuning scenarios can also be applied in our system before \textbf{Selective Parameter Encryption} and efficiently reducing the sizes of shared models directly helps with HE computation and communication efficiency (we also include preliminary results on this part in the appendix.

\subsubsection{Optimized Overheads}

\begin{figure}[t]
\includegraphics[width=0.45\textwidth]{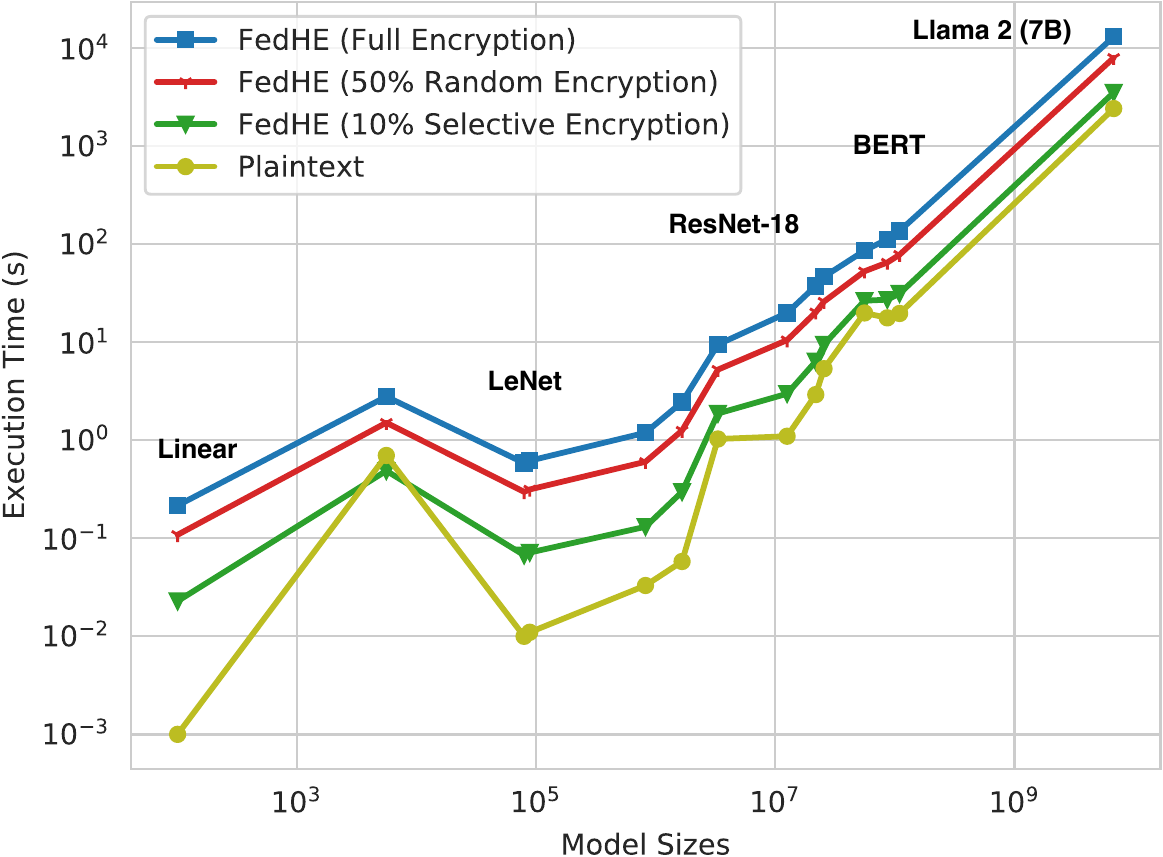}
\includegraphics[width=0.45\textwidth]{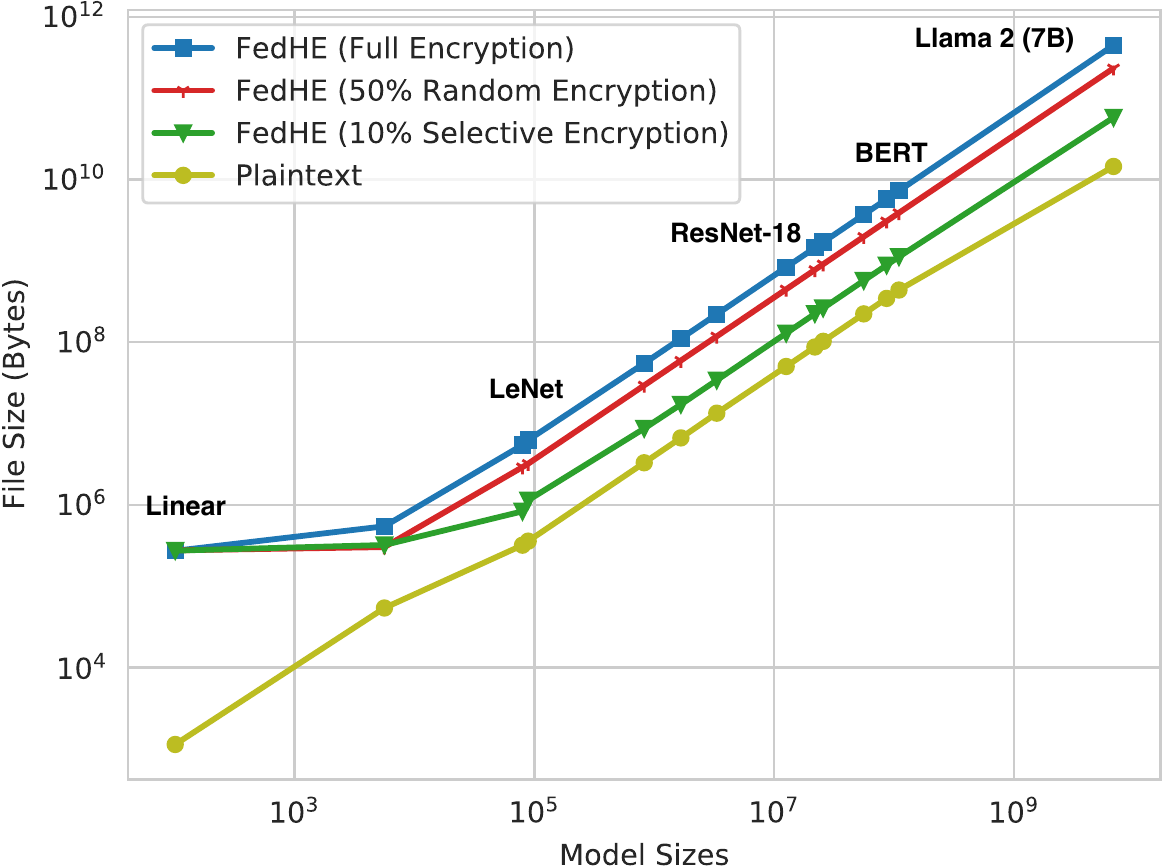}
\caption{Computational (up) and Computation (down) Overhead Comparison For Models of Different Sizes (logarithmic scale): 10\% Encryption is based on our selection strategy and 50\% encryption is based on random selection.}
\label{fig:comp_opt}
\end{figure}

We first examine the overhead optimization gains from \textbf{Selective Parameter Encryption}. We examine the overhead change when parameters with high privacy importance are selected and encrypted. Figure~\ref{fig:comp_opt} shows the overhead reduction from only encrypting certain parts of models, where both overheads are nearly proportional to the size of encrypted model parameters, which is coherent with the general relationship between HE overheads and input sizes. Note that after 10\% encryption per our \textbf{Selective Parameter Encryption}, the overheads are close to the ones of plaintext aggregation.  

\begin{figure*}[ht]

\centering
\includegraphics[width=0.333\textwidth]{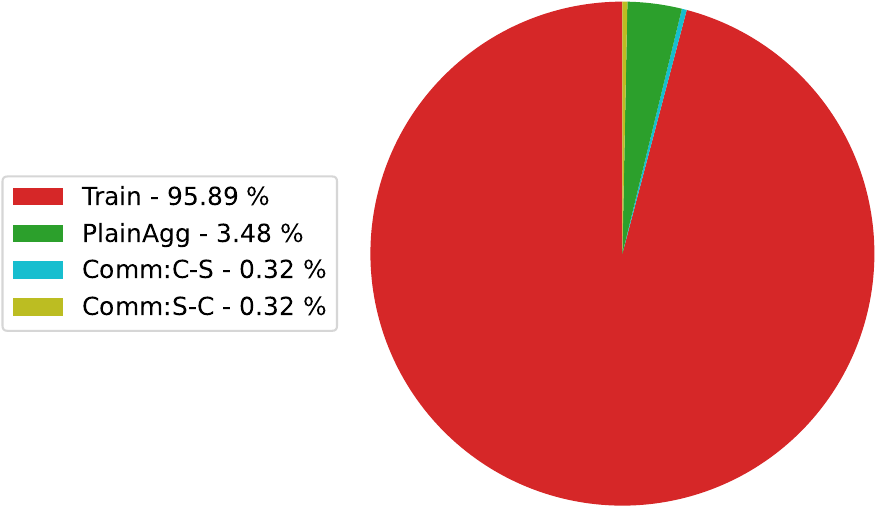}\hfill
\includegraphics[width=0.333\textwidth]{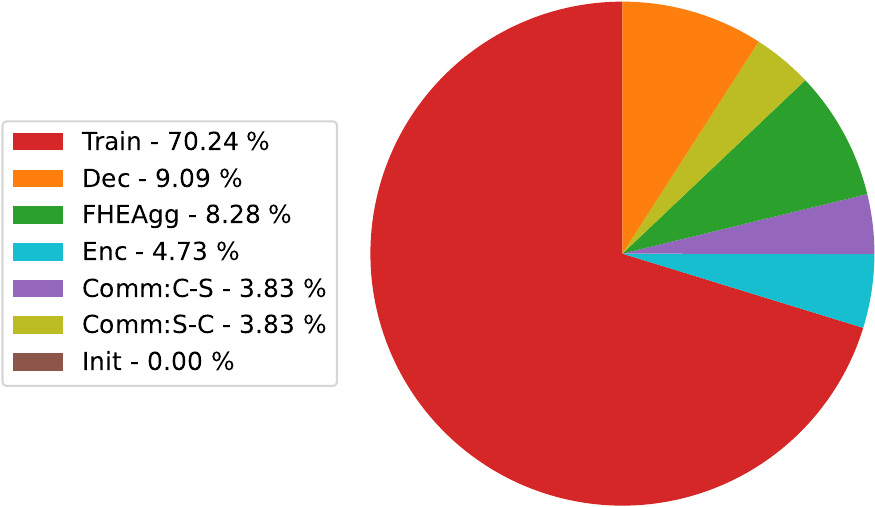}\hfill
\includegraphics[width=0.333\textwidth]{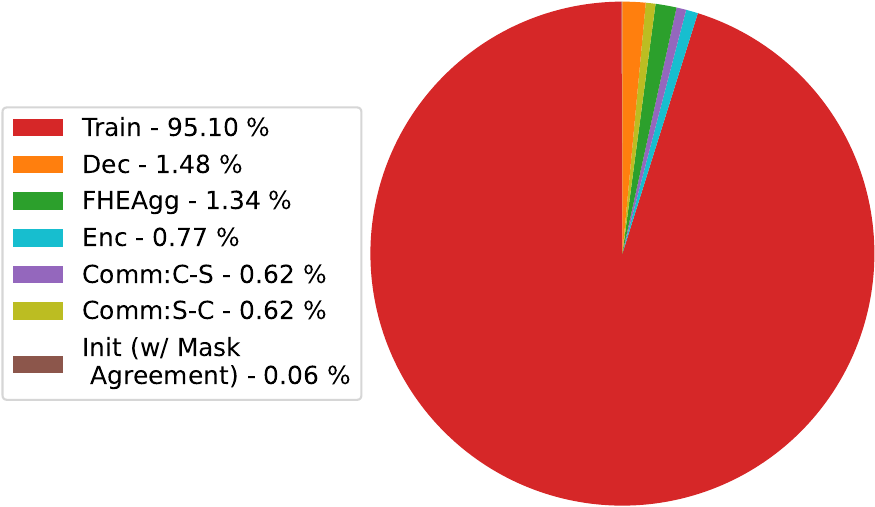}

\caption{Time Distribution of A Training Cycle on ResNet-50: with a single AWS region bandwidth of $200$ MB/s for plaintext FL (left), HE w/o optimization (middle), and HE w/ optimization (right). Optimization setup uses \textit{DoubleSqueeze}~\cite{tang2019doublesqueeze} with $k=1,000,000$ and encryption mask with an encrypted ratio $s = 30\%$.}
\label{fig:pie}
\end{figure*}

Figure~\ref{fig:pie} provides a perspective of overhead distribution to dissect the training cycle composition for the HE framework (both with and without optimizations) and the plaintext framework respectively with a single AWS region bandwidth. For a medium-sized model, the overheads (both computation and communication) from HE shift some portion of the local training procedure to aggregation-related steps compared to Non-HE, but not with an infeasible margin relatively speaking. Though generally smaller models require shorter training time, the overheads of the HE-based aggregation also drop proportionally.

\subsubsection{Effectiveness of Selection Defense}


\begin{figure*}[ht!]
\centering
\includegraphics[width=0.45\textwidth]{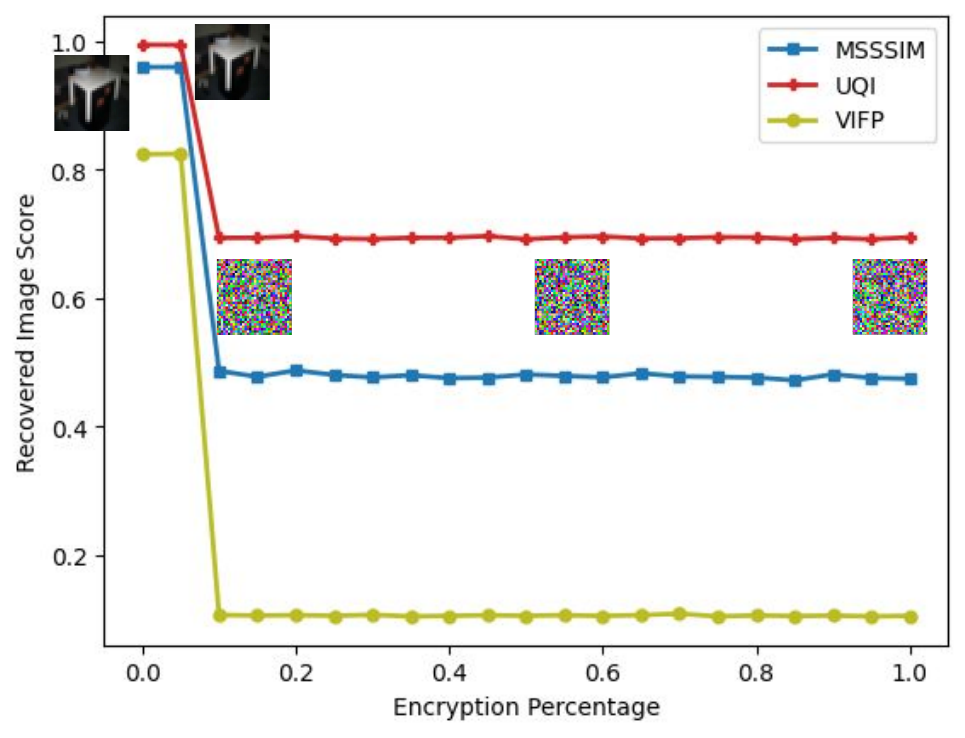}
\includegraphics[width=0.465\textwidth]{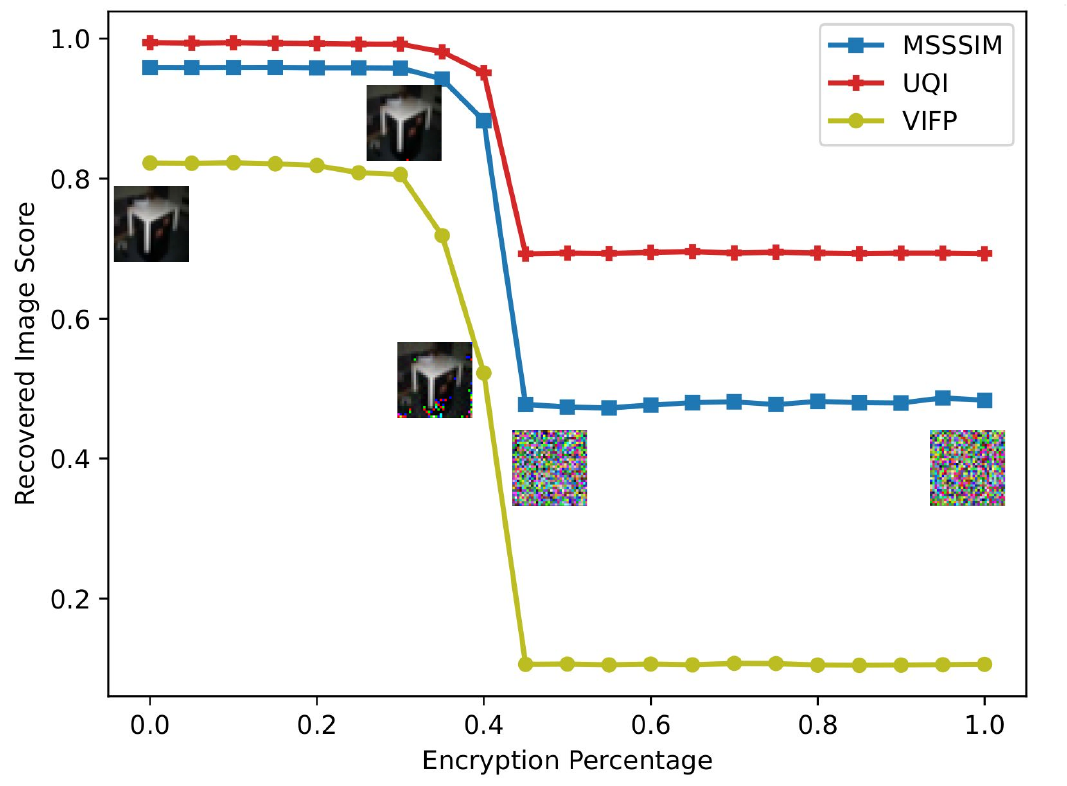}
\caption{Selection Protection Against Gradient Inversion Attack~\cite{zhu2019deep} On LeNet with the CIFAR-100 Dataset: attack results when protecting top-$s$ sensitive parameters (left) vs protecting random parameters (right). Each configuration is attacked 10 times and the best-recovered image is selected.}
\label{fig:attack_results}
\end{figure*}
\noindent To evaluate the defense effectiveness of \textbf{Selective Parameter Encryption}, we first use privacy sensitivity to generate a privacy map (Figure~\ref{fig:lenet_map}) and then verify the effectiveness of selection by performing gradient inversion (DLG~\cite{zhu2019deep}). We also provide defense results with Language Model Inversion Attacks~\cite{fowl2022decepticons} on Bert.

\underline{\textit{Defense effectiveness on CV tasks}}. We use image samples from CIFAR-100 to calculate the parameter sensitivities of the model. In the DLG attack experiments, we use Multi-scale Structural Similarity Index (MSSSIM), Visual Information Fidelity (VIF), and Universal Quality Image Index (UQI) as metrics to measure the similarity between recovered images and original training images to measure the attack quality hence the privacy leakage\footnote{The image similarity metric library used is at \url{https://pypi.org/project/sewar/}.}.
In Figure~\ref{fig:attack_results}, compared to random encryption selection where encrypting $42.5\%$ of the parameters can start to protect against attacks,  our top-$10\%$ encryption selection according to the model privacy map only alone can defend against the attacks, meaning lower overall overhead with the same amount of privacy protection.

\begin{figure*}[htp!]
\centering
\includegraphics[width=1\textwidth]{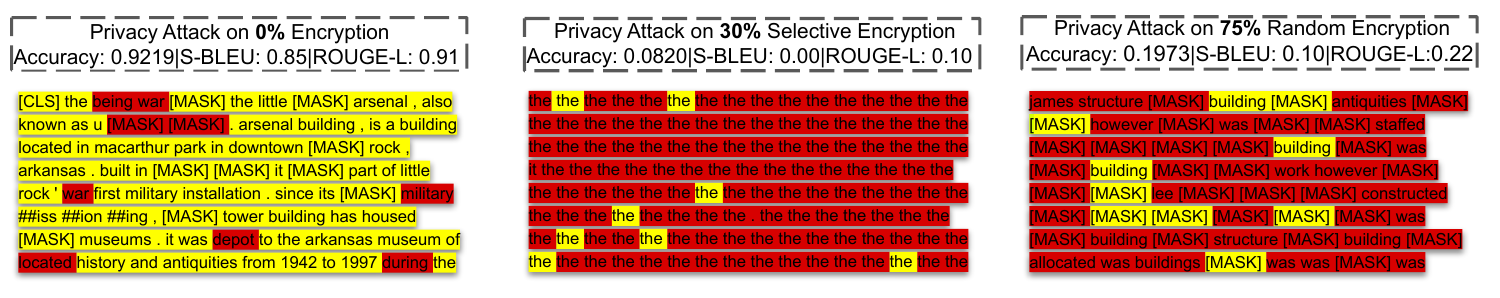}
\caption{Language Model Inversion Attacks~\cite{fowl2022decepticons} on Bert with the wikitext Dataset: \colorbox{red}{Red} indicates falsely-inverted words and \colorbox{yellow}{Yellow} indicates correctly-inverted words.}
\label{fig:bert-attack}
\end{figure*}

\underline{\textit{Defense effectiveness on NLP tasks}}. We use language samples from wikitext dataset in our experiment. As shown in Figure~\ref{fig:bert-attack}, with our sensitivity map indicating the top 30\% privacy-sensitive parameters, our encryption mask can prevent inversion attacks that yields better defense results than randomly encrypting 75\% of the model parameters.

\noindent\textbf{Empirical Selection Recipe.} Our selection strategy works by first encrypting more important model parameters. Empirically, from our experimental investigation, encrypting top-30\% most sensitive parameters, as well as the first and last model layers, tends to be robust to avoid information leakage~\cite{hatamizadeh2022gradvit} and attack defense (e.g. Figure~\ref{fig:lenet_map}), which can be used as a general guideline on top of model privacy maps.

%% file: background.tex
\section{Related Work}

\noindent\textbf{Existing Privacy Attacks On FL.}
Threats and attacks on privacy in the domain of Federated Learning have been studied in recent years~\cite{mothukuri2021survey}. General FL privacy attacks can be categorized into two types: inference attacks~\cite{nasr2019comprehensive, wang2019beyond, truex2019demystifying} and data leakage/reconstruction~\cite{criswell2014kcofi,bhowmick2018protection, hitaj2017deep}. Attacks are usually carried out on the models to retrieve certain properties of data providers or even reconstruct the data in the training datasets. With direct access to more fine-grained local models trained on a smaller dataset~\cite{wang2019beyond}, the adversary can have a higher chance of a successful attack.  Moreover, further attacks can be performed using GAN-based attacks to even fully recover the original data~\cite{hitaj2017deep}. The majority of the privacy attacks can be traced back to the direct exposure of plaintext accesses to local models to other parties (usually the server).

\noindent\textbf{Existing Non-HE Defense Mechanism.}
Local differential privacy has been adopted to protect local model updates by adding differential noise on the client side before the server-side aggregation ~\cite{truex2019hybrid,byrd2020differentially} where privacy guarantee requires large-scale statistical noise on fine-grained local updates that generally degrades model performance by a large margin. On the other hand, other work proposes to apply zero-sum masks (usually pair-wise) to mask local model updates such that any individual local update is indistinguishable to the server~\cite{bonawitz2017practical, so2022lightsecagg}. However, such a strategy introduces several challenges including key/mask synchronization requirements and federated learner dropouts. Compared to these solutions providing privacy protection in FL, HE is non-interactive and dropout-resilient (vs. general secure aggregation protocols~\cite{bonawitz2017practical, so2022lightsecagg}) and it introduces negligible model performance degradation (vs. noise-based differential privacy solutions~\cite{truex2019hybrid,byrd2020differentially}).

\noindent\textbf{Existing HE-based FL Work.}
Existing HE-based FL work either apply restricted HE schemes (e.g. additive scheme Paillier)~\cite{zhang2020batchcrypt,fang2021privacy, jiang2021flashe} without extensibility to further FL aggregation functions as well as sufficient performance and security guarantee (due to Paillier) or provide a generic HE implementation on FL aggregation~\cite{roth2022nvidia,ibmfl, jiang2021flashe, du2023efficient, ma2022privacy}. However, previous work still leaves the HE overhead increase issue as an open question. In our work, we propose a universal optimization scheme to largely reduce the overhead while providing promised privacy guarantees in a both systematic and algorithmic fashion, which makes HE-based FL viable in practical deployments. 



%% file: conclusion.tex
\section{Conclusion}
\noindent In this paper, we propose FedML-HE, the first practical Homomorphic Encryption-based privacy-preserving FL system that supports encryption key management, encrypted FL platform deployment, encryption optimizations to reduce overhead, and is designed to support efficient foundation model federated training. We design Selective Parameter Encryption that selectively encrypts the most privacy-sensitive parameters to minimize the size of encrypted model updates while providing customizable privacy preservation. 
Future work includes quantitative and theoretical analysis of the trade-offs among privacy guarantee, system overheads, and model performance compared to other approaches (including difference privacy and secure aggregation approaches), and improving threshold-HE's performance in the FL setting as well as supporting decentralized primitives such as Proxy Re-Encryption~\cite{ateniese2006improved}.

%% file: supp.tex
\newpage
\appendix
\section{Preliminaries}
\subsection{Federated Learning}
Federated learning is first proposed in~\cite{mcmahan2017communication}, which builds distributed machine learning models while keeping personal data on clients. Instead of uploading data to the server for centralized training, clients process their local data and share updated local models with the server. Model parameters from a large population of clients are aggregated by the server and combined to create an improved global model. 

The FedAvg~\cite{mcmahan2017communication} is commonly used on the server to combine client updates and produce a new global model. At each round, a global model $\mathbf{W}_\text {glob}$ is sent to $N$ client devices.
Each client $i$ performs gradient descent on its local data with $E$ local iterations to update the model $\mathbf{W}_i$.
The server then does a weighted aggregation of the local models to obtain a new global model, $\mathbf{W}_{\text {glob}}=\sum_{i=1}^N \alpha_i \mathbf{W}_i,$ where $\alpha_i$ is the weighting factor for client $i$.

Typically, the aggregation runs using plaintext model parameters through a central server  (in some cases, via a decentralized protocol), giving the server visibility of each local client's model in plaintext.

\subsection{Homomorphic Encryption}
\input{fhe.tex}
 Homomorphic Encryption is a cryptographic primitive that allows computation to be performed on encrypted data without revealing the underlying plaintext. It usually serves as a foundation for privacy-preserving outsourcing computing models. HE has generally four algorithms (\textit{KeyGen}, \textit{Enc}, \textit{Eval}, \textit{Dec}) as defined in Figure~\ref{fig:fhe}. The fundamental concept is to encrypt data prior to computation, perform the computation on the encrypted data without decryption, and then decrypt the resulting ciphertext to obtain the final plaintext.

Since FL model parameters are usually not integers, our method is built on the Cheon-Kim-Kim-Song (CKKS) scheme~\cite{cheon2017homomorphic}, a (leveled) HE variant that can work with approximate numbers.

\section{Key Management And Threshold HE}
\label{sec:key_management}
Our general system structure assumes the existence of a potentially compromised aggregation server, which performs the HE-based secure aggregation. Alongside this aggregation server, there also exists a trusted key authority server that generates and distributes HE keys and related crypto context files to authenticated parties (as described previously in Algorithm 1 in the main paper. We assume there is no collusion between these two servers.

Moreover, secure computation protocols for more decentralized settings without an aggregation server are also available using cryptographic primitives such as Threshold HE~\cite{aloufi2021computing}, Multi-Key HE~\cite{aloufi2021computing}, and Proxy Re-Encryption~\cite{ateniese2006improved, jin2022secure}. In such settings, secure computation and decryption can be collaboratively performed across multiple parties without the need for a centralized point. We plan to introduce a more decentralized version of FedML-HE in the future. Due to the collaborative nature of such secure computation, the key management will act more as a coordination point instead of a trusted source for key generation. 

\begin{figure}[ht]
\centering
\includegraphics[width=0.5\textwidth]{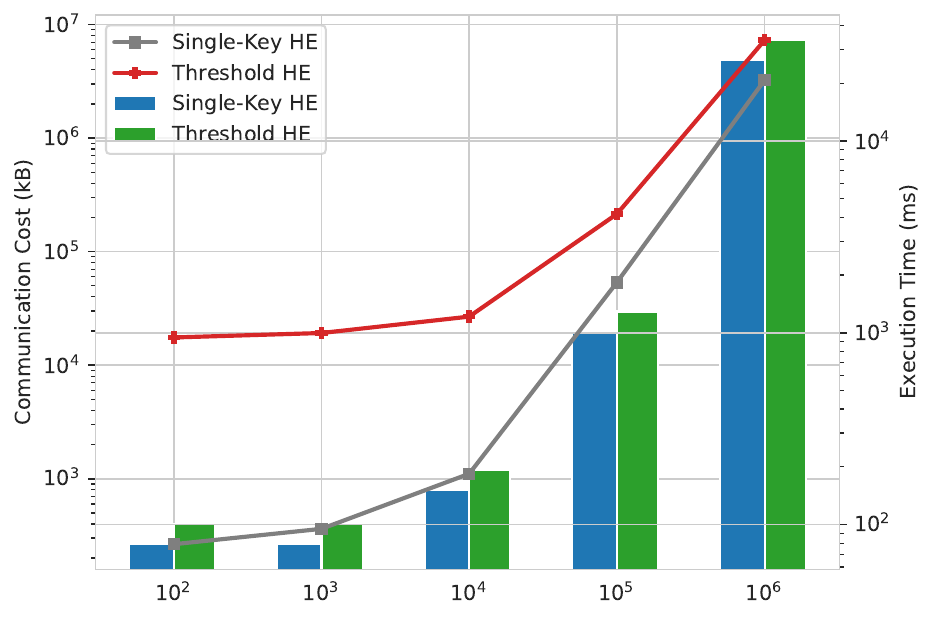}
\caption{Microbenchmark of Threshold-HE-Based FedAvg Implementation: we use a two-party threshold setup. Both the single-key variant and the threshold variant are configured with an estimated precision of 36 for a fair comparison.}
\label{fig:threshold-he}
\end{figure}

The threshold variant of HE schemes is generally based on Shamir's secret sharing~\cite{shamir1979share} (which is also implemented in PALISADE). Key generation/agreement and decryption processes are in an interactive fashion where each party shares partial responsibility of the task. Threshold key generation results in each party holding a share of the secret key and threshold decryption requires each party to partially decrypt the final ciphertext result and merge to get the final plaintext result. We provide benchmarkings of the threshold-HE-based FedAvg implementation in Figure~\ref{fig:threshold-he}.

\section{Framework APIs and Platform Deployment}
\subsection{Framework APIs}
Table~\ref{tab:apis} shows the framework APIs in our system related to HE.

\begin{table*}[ht!]
    \centering
    \begin{tabular}{|c|c|}
    \hline
        API Name & Description \\ \hline
        \textit{pk, sk} =  \textbf{key\_gen}(\textit{params}) & \makecell{Generate a pair of HE keys\\(public key and private key)} \\ \hline
        \textit{1d\_local\_model} = \textbf{flatten}(\textit{local\_model}) & \makecell{Flatten local trained model\\tensors into a 1D local model}\\ \hline
        \textit{enc\_local\_model} = \textbf{enc}(\textit{pk, 1d\_model}) & Encrypt the 1D model \\ \hline
        \makecell{\textit{enc\_global\_model} = \textbf{he\_aggregate}(\\\textit{enc\_models[n], weight\_factors[n]})} & \makecell{Homomorphically aggregate\\a list of 1D local models}\\ \hline
        \textit{dec\_global\_model} = \textbf{dec}(\textit{sk, enc\_global\_model}) & Decrypt the 1D global model \\ \hline
        \makecell{\textit{global\_model} = \textbf{reshape}(\\\textit{dec\_global\_model, model\_shape})} & \makecell{Reshape the 1D global model\\back to the original shape}\\ \hline
    \end{tabular}
\caption{HE Framework APIs}
\label{tab:apis}
\end{table*}

\subsection{Deploy Anywhere: A Deployment Platform MLOps For Edges/Cloud}

We implement our deployment-friendly platform such that FedML-HE can be easily deployed across cloud and edge devices.. Before the training starts, a user uploads the configured server package and the local client package to the web platform. The server package defines the operations on the FL server, such as the aggregation function and client sampling function; the local client package defines the customized model architecture to be trained (model files will be distributed to edge devices in the first round of the training). Both packages are written in Python. The platform then builds and runs the docker image with the uploaded server package to operate as the server for the training with edge devices configured using the client package.

\begin{figure*}[ht]
\centering
\includegraphics[width=0.9\textwidth]{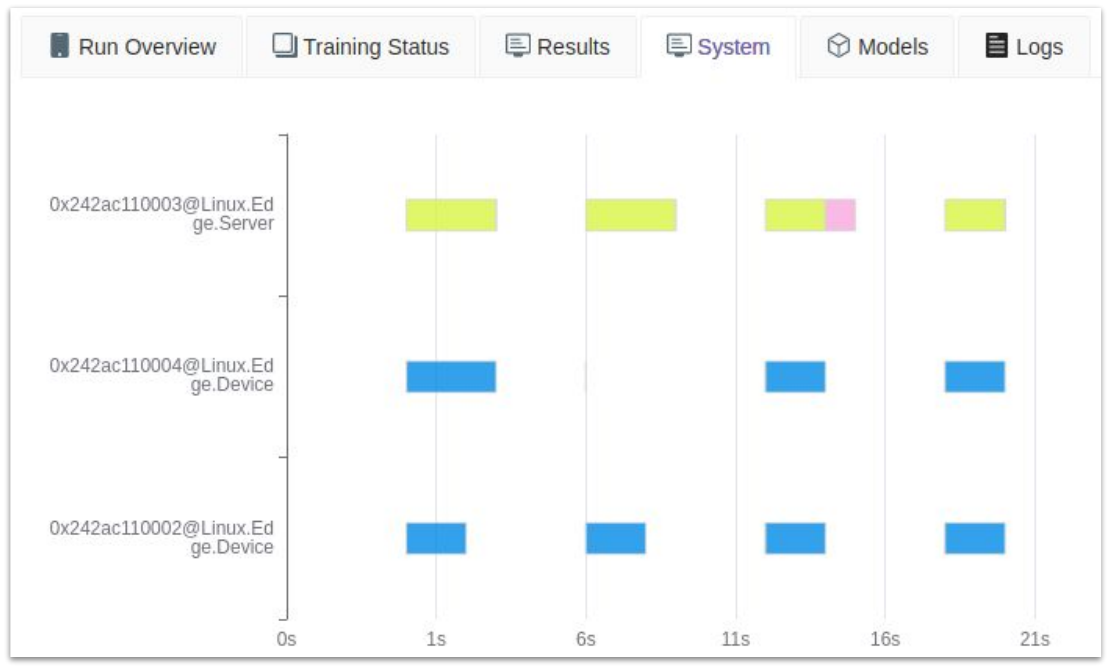}
\caption{Deployment Interface Example of FedML-HE: Overhead distribution monitoring on each edge device (e.g. Desktop (Ubuntu), Laptop (MacBook), and Raspberry Pi 4), which can be used to pinpoint HE overhead bottlenecks and guide optimization.}
\label{fig:mlops}
\end{figure*}

As shown in Figure~\ref{fig:mlops}, during the training, users can also keep tracking the learning procedure including device status, training progress/model performance, and FedML-HE system overheads (e.g., training time, communication time, CPU/GPU utilization, and memory utilization) via the web interface. Our platform keeps close track of overheads, which allows users to in real-time pinpoint HE overhead bottlenecks if any.

\section{Additional Experiments}
\label{sec:he-benchmark}

\begin{table*}[ht]
    \small
    \centering
    \begin{tabular}{|c|c|c|c|c|c|c|c|}
    \hline\hline
        Model & Model Size & \makecell[l]{HE\\ Time (s)} & \makecell[l]{Non-HE\\ Time (s)} & \makecell[l]{Comp\\Ratio} & Ciphertext & Plaintext& \makecell[l]{Comm\\Ratio}  \\ \hline\hline
        Linear Model  & 101 & 0.216 & 0.001 & 150.85 & 266.00 KB & 1.10 KB& 240.83 \\ \hline
        \makecell[l]{TimeSeries\\Transformer} & 5,609 & 2.792 & 0.233 & 12.00 & 532.00 KB & 52.65 KB& 10.10 \\ \hline
        MLP (2 FC) & 79,510 & 0.586 & 0.010 & 60.46 & 5.20 MB& 311.98 KB & 17.05 \\ \hline
        LeNet  & 88,648 & 0.619 & 0.011 & 57.95 & 5.97 MB& 349.52 KB & 17.50 \\ \hline
        \makecell{RNN(2 LSTM\\ + 1 FC)}  & 822,570 & 1.195 & 0.013 & 91.82 & 52.47 MB& 3.14 MB& 16.70 \\ \hline
        \makecell{CNN (2 Conv\\ + 2 FC)} & 1,663,370 & 2.456 & 0.058 & 42.23 & 103.15 MB& 6.35 MB& 16.66 \\ \hline
        MobileNet  & 3,315,428 & 9.481 & 1.031 & 9.20 & 210.41 MB& 12.79 MB& 16.45 \\ \hline
        ResNet-18 & 12,556,426 & 19.950 & 1.100 & 18.14 & 796.70 MB& 47.98 MB& 16.61 \\ \hline
        ResNet-34 & 21,797,672 & 37.555 & 2.925 & 12.84 & 1.35 GB& 83.28 MB& 16.60 \\ \hline
        ResNet-50 & 25,557,032 & 46.672 & 5.379 & 8.68 & 1.58 GB& 97.79 MB& 16.58 \\ \hline
        GroupViT & 55,726,609 & 86.098 & 19.921 & 4.32 & 3.45 GB& 212.83 MB& 16.61 \\ \hline
        \makecell{Vision\\ Transformer} & 86,389,248 & 112.504 & 17.739 & 6.34 & 5.35 GB& 329.62 MB&16.62 \\ \hline
        BERT & 109,482,240 & 136.914 & 19.674 & 6.96 & 6.78 GB& 417.72 MB& 16.62 \\ \hline
        Llama 2 & 6.74 B  & 13067.154 & 2423.976 & 5.39 & 417.43 GB& 13.5 GB& 30.92 \\ \hline
    \end{tabular}
    \caption{Vanilla Fully-Encrypted Models of Different Sizes: with $3$ clients; Comp Ratio is calculated by time costs of HE over time costs of Non-HE; Comm Ratio is calculated by file sizes of HE over file sizes of Non-HE. CKKS is configured with default crypto parameters.}
    \label{tab:models}
\end{table*}


We evaluate the HE-based training overheads (without our optimization in place) across various FL training scenarios and configurations. This analysis covers diverse model scales, HE cryptographic parameter configurations, client quantities involved in the task, and communication bandwidths. This helps us to identify bottlenecks in the HE process throughout the entire training cycle. We also benchmark our framework against other open-source HE solutions to demonstrate its advantages.

\subsection{Parameter Efficiency Techniques in HE-Based FL}
\label{sec:eff_exps}
Table~\ref{tab:eff} shows the optimization gains by applying model parameter efficiency solutions in HE-Based FL.

\begin{table}[ht]
    \centering
    \begin{tabular}{|c|c|c|c|}
    \hline
        \makecell{Models} & PT (MB)& CT & \makecell{Opt\\(MB)} \\ \hline
        \makecell{ResNet-18 \\(12 M)\\~\cite{tang2019doublesqueeze}} & 47.98 & 796.70 MB & 19.03  \\\hline
        \makecell{BERT\\ (110 M)\\~\cite{hu2021lora}} & 417.72 & 6.78 GB & 16.66  \\\hline
    \end{tabular}
    \caption{Parameter Efficiency Overhead: PT means plaintext and CT means ciphertext. Communication reductions are 0.60 and 0.96.}
    \label{tab:eff}
\end{table}

\subsection{Results on Different Scales of Models}
\label{sec:models}
\hfill\\
We evaluate our framework on models with different size scales and different domains, from small models like the linear model to large foundation models such as Vision Transformer~\cite{dosovitskiy2020image} and BERT~\cite{devlin2018bert}. As Table~\ref{tab:models} show, both computational and communicational overheads are generally proportional to model sizes.

Table~\ref{tab:models} illustrates more clearly the overhead increase from the plaintext federated aggregation. The computation fold ratio is in general $5$x $\sim 20$x while the communication overhead can jump to a common $15$x. Small models tend to have a higher computational overhead ratio increase. This is mainly due to the standard HE initialization process, which plays a more significant role when compared to the plaintext cost. The communication cost increase is significant for models with sizes smaller than $4096$ (the packing batch size) numbers. Recall that the way our HE core packs encrypted numbers makes an array whose size is smaller than the packing batch size still requires a full ciphertext.

\begin{table}[ht!]
    \centering
    \begin{tabular}{|c|c|c|c|c|}
    \hline\hline
        \makecell{HE\\Batch\\Size} & \makecell{Scaling\\Bits} & \makecell{Comp\\(s)} & \makecell{Comm\\(MB)} & \makecell{Model Test \\Accuracy \\$\Delta$ (\%)} \\ 
        \hline\hline
        1024 & 14 & 8.834 & 407.47 & -0.28 \\ \hline
        1024 & 20 & 7.524 & 407.47 & -0.21 \\ \hline
        1024 & 33 & 7.536 & 407.47 & 0 \\ \hline
        1024 & 40 & 7.765 & 407.47 & 0 \\ \hline
        1024 & 52 & 7.827 & 407.47 & 0 \\ \hline
        2048 & 14 & 3.449 & 204.50 & -0.06 \\ \hline
        2048 & 20 & 3.414 & 204.50 & -0.13 \\ \hline
        2048 & 33 & 3.499 & 204.50 & 0 \\ \hline
        2048 & 40 & 3.621 & 204.50 & 0 \\ \hline
        2048 & 52 & 3.676 & 204.50 & 0 \\ \hline
        4096 & 14 & 1.837 & 103.15 & -1.85 \\ \hline
        4096 & 20 & 1.819 & 103.15 & 0.32 \\ \hline
        4096 & 33 & 1.886 & 103.15 & 0 \\ \hline
        4096 & 40 & 1.998 & 103.15 & 0 \\ \hline
        4096 & 52 & 1.926 & 103.15 & 0 \\ \hline
    \end{tabular}
\caption{Computational \& Communicational Overhead of Different Crypto Parameter Setups: tested with CNN (2 Conv+ 2 FC) and on 3 clients; model test accuracy $\Delta$s is the difference between the best plaintext global model and the best global encrypted global models.}
\vspace{-0.5cm}
\label{tab:params}
\end{table}

\subsection{Results on Different Cryptographic Parameters}
\hfill\\
We evaluate the impacts of variously-configured cryptographic parameters. We primarily look into the packing batch size and the scaling bits. The packing batch size determines the number of slots packed in a single ciphertext while the scaling bit number affects the ``accuracy'' (i.e., how close the decrypted ciphertext result is to the plaintext result) of approximate numbers represented from integers. 

From Table~\ref{tab:params}, the large packing batch sizes in general result in faster computation speeds and smaller overall ciphertext files attributed to the packing mechanism for more efficiency. However, the scaling factor number has an almost negligible impact on overheads.

Unsurprisingly, it aligns with the intuition that the higher bit scaling number results in higher ``accuracy'' of the decrypted ciphertext value, which generally means the encrypted aggregated model would have a close model test performance to the plaintext aggregated model. However, it is worth mentioning that since CKKS is an approximate scheme with noises, the decrypted aggregated model can yield either positive or negative model test accuracy $\Delta$s, but usually with a negative or nearly zero $\Delta$.

\subsection{Impact from Number of Clients}
\begin{figure}[htbp]
  \centering

  \begin{subfigure}[t]{0.48\textwidth}
    \centering
    \includegraphics[width=\textwidth]{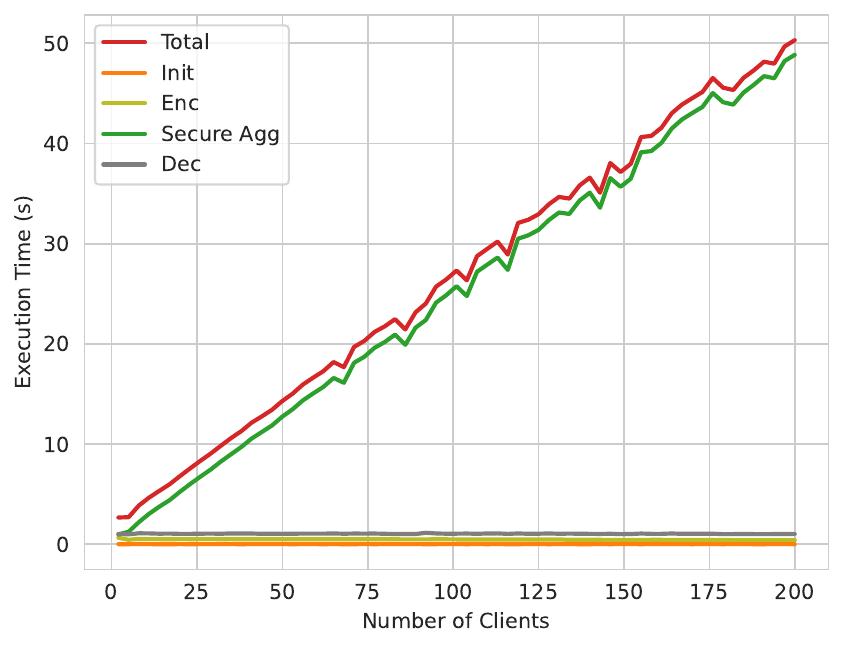}
    \caption{Step Breakdown of HE Computational Cost vs. Number of Clients (Up to 200): tested on fully-encrypted CNN}
    \label{fig:n_200}
  \end{subfigure}
  \hfill
  \begin{subfigure}[t]{0.48\textwidth}
    \centering
        \includegraphics[width=\textwidth]{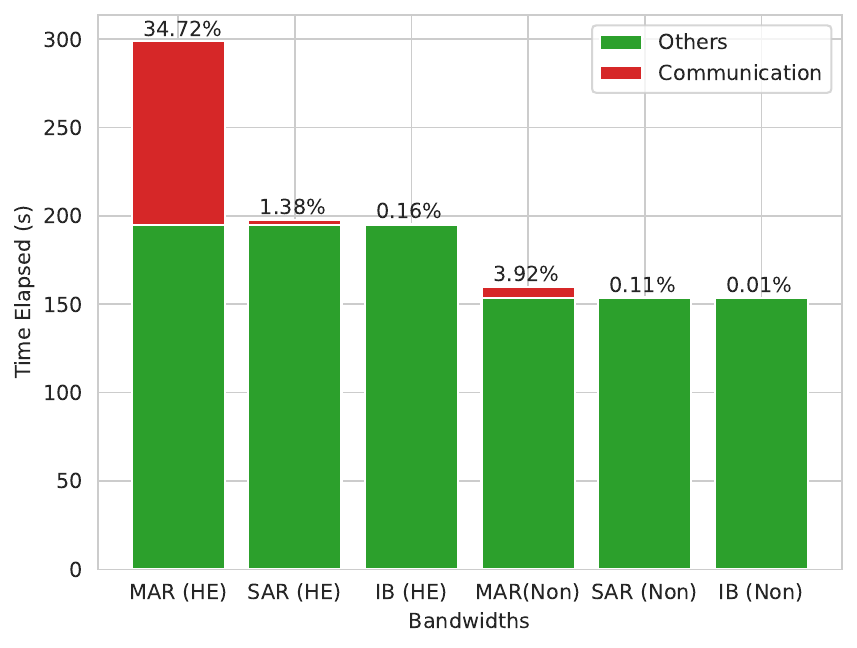}
        \caption{Impact of Different Bandwidths on Communication and Training Cycles on Fully-Encrypted ResNet-50: HE means HE-enabled training and Non means plaintext. Others include all other procedures except communication during training. Percentages represent the portion of communication cost in the entire training cycle.}
        \label{fig:comm_bar}
    \end{subfigure}
    \caption{Results on Different Number of Clients and Communication Setup}
\end{figure}

\hfill\\
As real-world systems often experience a dynamic amount of participants within the FL system, we evaluate the overhead shift over the change in the number of clients. Figure~\ref{fig:n_200} breaks down the cost distribution as the number of clients increases. With a growing number of clients, it also means proportionally-added ciphertexts as inputs to the secure aggregation function thus the major impact is cast on the server. When the server is overloaded, our system also supports client selection to remove certain clients without largely degrading model performance.

\subsection{Communication Cost on Different Bandwidths}
FL parties can be allocated in different geo-locations which might result in communication bottlenecks. Typically, there are two common scenarios: (inter) data centers and (intra) data centers. In this part, we evaluate the impact of the bandwidths on communication costs and how it affects the FL training cycle. We categorize communication bandwidths using $3$ cases:
\begin{itemize}
    \item Infiniband (IB): communication between intra-center parties. 5 GB/s as the test bandwidth.
    \item Single AWS Region (SAR): communication between inter-center parties but within the same geo-region (within US-WEST). 592 MB/s as the test bandwidth.
        \item Multiple AWS Region (MAR): communication between inter-center parties but across the different geo-region (between US-WEST and EU-NORTH). 15.6 MB/s as the test bandwidth.
\end{itemize}

As shown in Figure~\ref{fig:comm_bar}, we deploy FedML-HE on $3$ different geo-distributed environments, which are operated under different bandwidths. It is obvious that the secure HE functionality has an enormous impact on low-bandwidth environments while medium-to-high-bandwidth environments suffer limited impact from increased communication overhead during training cycles, compared to Non-HE settings.

\subsection{Different Encryption Selections}
Table~\ref{tab:selection} shows the overhead reductions with different selective encryption rates.

\begin{table}[h]
    \centering
    \begin{tabular}{|c|c|c|c|c|}
    \hline
        Selection & \makecell{Comp\\(s)} & Comm & \makecell{Comp\\Ratio} & \makecell{Comm\\Ratio} \\ \hline\hline
        
        Enc w/ 0\% & 17.739 & 329.62 MB & 1.00 & 1.00 \\ \hline
        Enc w/ 10\% & 30.874 & 844.49 MB & 1.74 & 2.56 \\ \hline
        Enc w/ 30\%  & 50.284 & 1.83 GB & 2.83 & 5.69 \\ \hline
        Enc w/ 50\% & 70.167 & 2.83 GB & 3.96 & 8.81 \\ \hline
        Enc w/ 70\%& 88.904 & 3.84 GB & 5.01 & 11.93 \\ \hline
        Enc w/ All & 112.504 & 5.35 GB & 6.34 & 16.62 \\ \hline
    \end{tabular}
    \caption{Overheads With Different Parameter Selection Configs Tested on Vision Transformer: ``Enc w/ 10\%'' means performs encrypted computation only on 10\% of the parameters; all computation and communication results include overheads from plaintext aggregation for the rest of the parameters.}
    \label{tab:selection}
\end{table}

\subsection{Comparison with Other FL-HE Frameworks}
\label{sec:diff_frameworks}

\begin{table*}
    \centering
    \begin{tabular}{|c|c|c|c|c|c|}
    \hline\hline
        Frameworks &HE Core& \makecell{Key\\Management}& Comp (s) & \makecell{Comm \\(MB)} & \makecell{HE\\Multi-Party\\ Functionalities}\\ \hline
        \hline
        Ours &PALISADE & \makecell{\cmark}  & 2.456 & 105.72  & \makecell{PRE,\\ ThHE}\\
                \hline
        Ours (w/ Opt) &PALISADE & \makecell{\cmark} & 0.874 & 16.37  & \makecell{PRE,\\ ThHE}\\
                \hline    
        Ours & \makecell{SEAL\\(TenSEAL)} & \makecell{\cmark}  & 3.989 & 129.75  & ---\\\hline
        \makecell{Nvidia FLARE\\(9a1b226)} &\makecell{SEAL\\(TenSEAL)}& \makecell{\cmark} & 2.826 & 129.75  &---\\ \hline
        \makecell{IBMFL\\(8c8ab11)}&\makecell{SEAL\\(HELayers)} & $\bigcirc$ & 3.955 & 86.58  & ---\\ \hline
        
        Plaintext&--- & --- & 0.058 & 6.35  & ---\\ \hline
    \end{tabular}
    \caption{Different Frameworks: tested with CNN (2 Conv
+ 2 FC) and on 3 clients; Github commit IDs are specified. For key management, our work uses a key authority server; FLARE uses a security content manager; IBMFL currently provides a local simulator.}
    \label{tab:frameworks}
\end{table*}

We compare our framework to the other open-sourced FL frameworks with HE capability, namely NVIDIA FLARE (NVIDIA) and IBMFL. 

Both NVIDIA and IBMFL utilize Microsoft SEAL as the underlying HE core, with NVIDIA using OpenMinded's python tensor wrapper over SEAL and TenSEAL; IBMFL using IBM'spython wrapper over SEAL and HELayers (HELayers also has an HElib version). Our HE core module can be replaced with different available HE cores, to give a more comprehensive comparison, we also implement a TenSEAL version of our framework for evaluation.

Table~\ref{tab:frameworks} demonstrates the performance summary of different FedML-HE frameworks using an example of a CNN model with $3$ clients. Our PALISADE-powered framework has the smallest computational overhead due to the performance of the PALISADE library. In terms of communication cost, FedML-HE (PALISADE) comes second after IBMFL's smallest file serialization results due to the efficient packing of HELayers' Tile tensors~\cite{aharoni2011helayers}. 

Note that NVIDIA's TenSEAL-based realization is faster than the TenSEAL variant of our system. This is because NVIDIA scales each learner's local model parameters locally rather than weighing ciphertexts on the server. This approach reduces the need for the one multiplication operation usually performed during secure aggregation (recall that HE multiplications are expensive). However, such a setup would not suit the scenario where the central server does not want to reveal its weighing mechanism per each individual local model to learners as it reveals partial (even full in some cases) information about participants in the system.

%% file: fhe.tex
\begin{figure}[ht]
\begin{mdframed}

\begin{itemize}
\item HE.\textit{KeyGen}($\lambda$): given the security parameter $\lambda$, the key generation algorithm outputs a key pair $(pk, sk)$ and the related cryptographic context.
    
\item HE.\textit{Enc}($pk, m$):the encryption algorithm takes in $pk$ and a plaintext message $m$, then
outputs the ciphertext $c$.

\item HE.\textit{Eval}($c, f$):the encrypted evaluation algorithm takes in a ciphertext message $c$ and a function $f$, then outputs the computation result $c'$.

\item HE.\textit{Dec}($sk, c'$):the encryption algorithm takes in $sk$ and a ciphertext message $c'$, then
outputs the plaintext $m'$.

\end{itemize}
\end{mdframed}
\caption{General Scheme of Homomorphic Encryption}
\label{fig:fhe}
\end{figure}